\PassOptionsToPackage{table}{xcolor}
\documentclass{article}
\usepackage{iclr2027_conference,times}
\usepackage{etoolbox}
\makeatletter
\patchcmd{\@maketitle}
  {\lhead{Published as a conference paper at ICLR 2027}}
  {\lhead{}}
  {}{}
\patchcmd{\@maketitle}
  {{\LARGE\sc \@title\par}}
  {{\centering\LARGE\normalfont \@title\par}}
  {}{}
\patchcmd{\@maketitle}
  {\begin{tabular}[t]{l}\bf\rule{\z@}{24pt}\@author\end{tabular}}
  {\hfil\begin{tabular}[t]{c}\bf\rule{\z@}{24pt}\@author\end{tabular}\hfil}
  {}{}
\makeatother

\renewcommand{\headrulewidth}{0pt}

\usepackage{amsmath,amsfonts,bm}

\def\eqref#1{equation~\ref{#1}}

\def\1{\bm{1}}

\DeclareMathAlphabet{\mathsfit}{\encodingdefault}{\sfdefault}{m}{sl}
\SetMathAlphabet{\mathsfit}{bold}{\encodingdefault}{\sfdefault}{bx}{n}

\usepackage{xcolor}
\definecolor{SparsePRBlue}{HTML}{0057B8}
\makeatletter
\def\section{\@startsection {section}{1}{\z@}{-2.0ex plus
    -0.5ex minus -.2ex}{1.5ex plus 0.3ex
minus0.2ex}{\large\sc\raggedright\color{SparsePRBlue}}}
\def\subsection{\@startsection{subsection}{2}{\z@}{-1.8ex plus
-0.5ex minus -.2ex}{0.8ex plus .2ex}{\normalsize\sc\raggedright\color{SparsePRBlue}}}
\def\subsubsection{\@startsection{subsubsection}{3}{\z@}{-1.5ex
plus -0.5ex minus -.2ex}{0.5ex plus .2ex}{\normalsize\sc\raggedright\color{SparsePRBlue}}}
\makeatother
\usepackage{hyperref}
\hypersetup{
  colorlinks=true,
  linkcolor=SparsePRBlue,
  citecolor=SparsePRBlue,
  urlcolor=SparsePRBlue
}
\usepackage{url}
\usepackage{graphicx}
\usepackage{comment}
\usepackage{wrapfig}
\usepackage{algorithm}
\usepackage{algpseudocode}
\usepackage{amssymb}
\usepackage{amsmath} 
\usepackage{booktabs}
\usepackage{multirow}
\usepackage{makecell}
\usepackage{adjustbox}
\usepackage{graphicx}

\usepackage{newfloat}
\usepackage{listings}
\usepackage{caption}

\usepackage{placeins}
\usepackage{dblfloatfix}
\usepackage{tabularx}
\usepackage{array}

\definecolor{NVIDIAGreen}{HTML}{76B900}
\definecolor{NVIDIALight}{HTML}{EBF5FF}
\definecolor{NVIDIAGreen}{HTML}{76B900}
\definecolor{NVIDIAMedium}{HTML}{E4F1CC}  
\definecolor{NVIDIADark}{HTML}{1E1E1E}    

\title{Partition the Support, Reconstruct the \\ Residual: Training-Free Sparse Attention for Video Generation and World Models}

\author{
{\normalfont\large Pardis Taghavi, Reza Langari, Gaurav Pandey}\\[1ex]
{\normalfont Texas A\&M University}\\
{\normalfont\ttfamily \{ptgh,rlangari,gpandey\}@tamu.edu}
}

\iclrfinalcopy 

\begin{document}
\addtocontents{toc}{\protect\setcounter{tocdepth}{-1}}

\maketitle
\raggedbottom
\setlength{\textfloatsep}{18pt}

\begin{abstract}

Block-sparse attention accelerates video transformers by grouping queries and keys and computing attention only between selected groups. However, it faces two challenges: (1) queries sharing a block route may attend to different keys; and (2) retaining most attention mass does not necessarily preserve the attention output. We find that partition choice affects both the overlap among grouped queries’ preferred supports and how well an affine function of the sparse output can represent the difference between dense and sparse outputs. We introduce SparsePR, a training-free method combining Response-Coupled Partitioning with Probe-Fitted Residual Reconstruction. We first group paired key/value tokens by how their keys respond to sampled queries. We then group queries by their responses to the resulting key-group centroids for shared sparse routing. We compute exact attention on a small set of query rows and fit an affine correction to the differences between their dense and sparse outputs using weighted ridge regression. The fitted map predicts a residual for each unprobed query, which we add to its sparse output. We measure attention-output error as the relative \(L_2\) error with respect to each query's dense output. Across four video generation and world models, SparsePR reduces this error, averaged over unprobed queries, by 61.6--90.5\% compared with semantic block-sparse attention without residual correction, at 22\% executed-pair density including exact probes. Ablations show that response-coupled partitioning reduces error before correction and improves reconstruction with the same probe budget, while probe fitting provides the largest overall reduction. SparsePR achieves \(1.48\times\)--\(2.61\times\) end-to-end speedups over dense attention, with dense-reference PSNR of 24.42--31.84 dB at 21.9--26.0\% realized executed-pair density.


\href{https://pardistaghavi.github.io/SparsePR-website/}{Project page}
\end{abstract}

\section{Introduction}
\label{sec:intro}

Video generation models~\citep{kong2024hunyuanvideo,wan2025,yang2024cogvideox}
and video world models~\citep{ali2025world,nvidia2026cosmos3}
process long spatiotemporal token sequences, making quadratic self-attention a major inference bottleneck. Attention often concentrates on a subset of tokens~\citep{zhang2025fast,chen2026sparse}, which training-free methods exploit by selectively computing
query--key interactions~\citep{svg,zhang2025spargeattention,luo2026attention}. For hardware efficiency, block-sparse methods organize queries and keys into blocks and compute selected block pairs.
These blocks can be organized by spatial and temporal
position~\citep{zhang2025fast,svg} or be formed by separately clustering query and key token
representations by semantic
similarity~\citep{svg2}. Queries within a group share the same selected key/value groups, even when individual queries may favor different keys. This shared selection can therefore omit interactions important
to some queries and change their attention outputs. Grouping queries with similar attention patterns can reduce this mismatch. However, skipped interactions can still cause the sparse output to differ from the dense output. Estimating this difference and adding it to the sparse output can improve accuracy without computing every interaction. 

We make three observations that further motivate our approach. First, individual queries may concentrate attention on small but different sets of keys. Per-query concentration alone therefore does not describe the overlap among these sets within a query group. Second, retained attention mass alone does not determine attention-output error. The error also depends on the value vectors associated with the skipped interactions. Similar retained mass can therefore produce different output errors. Third, partition choice affects how much of the difference between dense and sparse attention outputs can be represented as an affine function of the sparse output, a linear map with an offset. We assess this representability through the best fit using dense outputs on all evaluated queries. Prediction from a small probe set is evaluated separately. Together, these observations motivate designing token partitions that group queries with overlapping attention supports and make the resulting output error easier to approximate with an affine function of the sparse output.

\begin{figure*}[!t]
    \centering
    \includegraphics[width=0.99\linewidth]{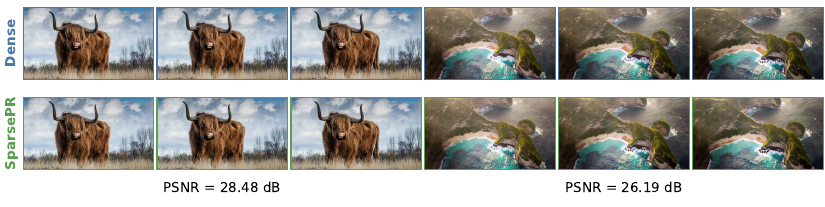}
    \caption{Qualitative results for the Cosmos-Predict2.5-14B model.}
    \label{fig:qualitative}
\end{figure*}

We introduce SparsePR, a training-free sparse-attention framework
that combines \emph{Response-Coupled Partitioning} and
\emph{Probe-Fitted Residual Reconstruction}.
Response-Coupled Partitioning groups paired key/value tokens
using low-dimensional coordinates derived from their keys'
responses to sampled queries. The resulting key group centroids define response coordinates for grouping queries that share a sparse route. We construct the K/V groups first, then the query groups without alternating between the two partitions in a single pass. Probe-Fitted Residual Reconstruction computes exact attention
for a small set of query rows and measures the differences
between their dense and sparse outputs. We use weighted ridge regression to fit an affine map from the sparse outputs to these differences, then apply it to unprobed queries and add the predicted corrections to their sparse outputs. Both components operate online and require no offline sparsity profiling.


We make three contributions. First, we examine how token partitioning affects shared attention support and the output error that an affine correction can represent, and show why retained attention mass alone does not determine attention output accuracy.
Second, we introduce SparsePR, a training-free method combining response-based token grouping with correction of
sparse attention outputs. Third, across four video generation and world models, SparsePR reduces mean relative attention output error on unprobed queries
by 61.6--90.5\% compared with semantic block-sparse attention without correction at 22\% executed-pair density including probes, and achieves $1.48\times$--$2.61\times$ end-to-end speedups over dense attention at 21.9--26.0\% realized executed-pair density.

\begin{figure}[!t]
    \centering
    \includegraphics[width=\textwidth]{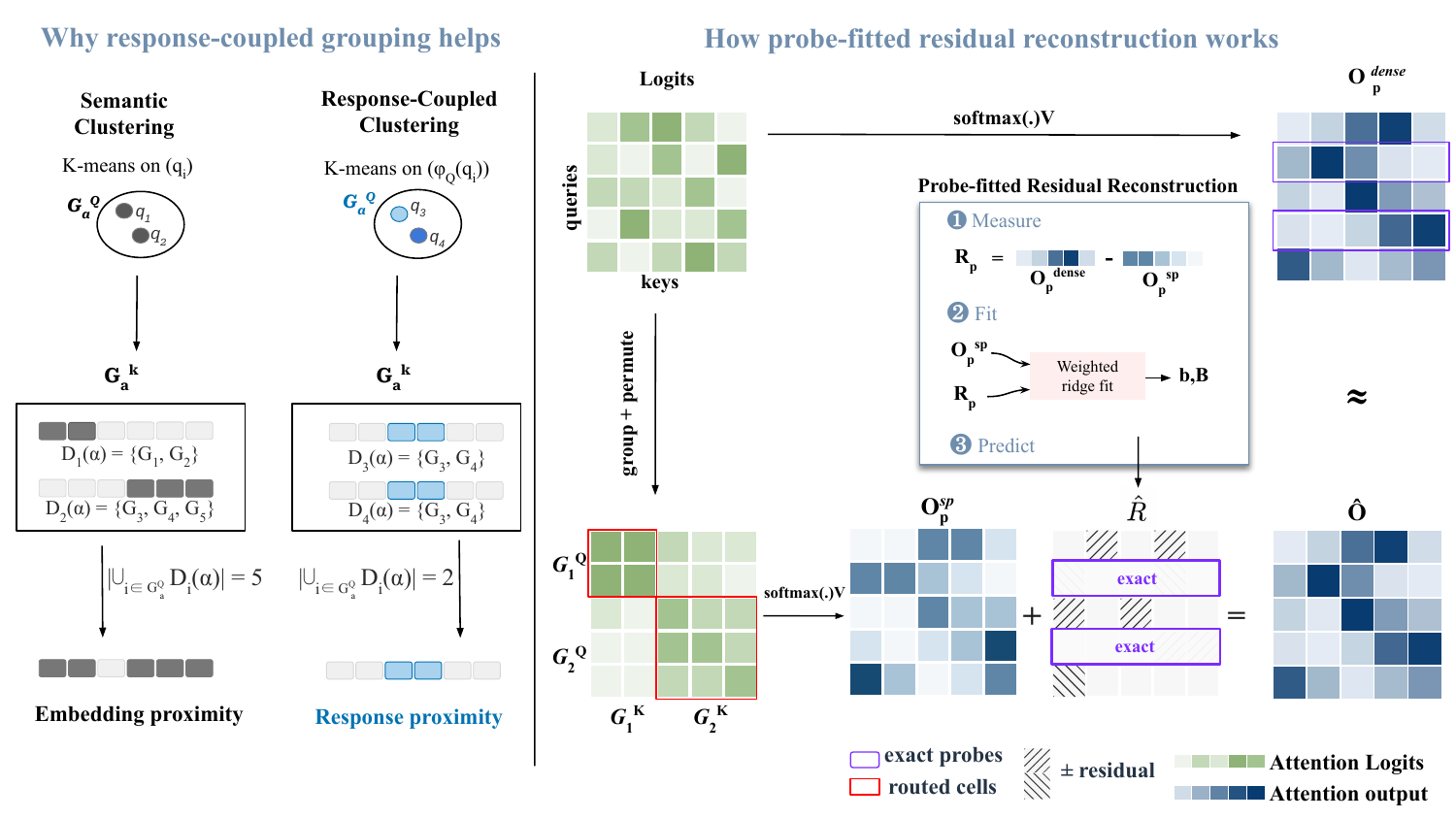}
    \caption{
Overview of SparsePR. Response-Coupled Partitioning builds executable K/V and query groups, and Probe-Fitted Residual Reconstruction uses exact probe rows to correct the sparse output.
}
    \label{fig:main}
    \vspace{-4pt}
\end{figure}

\section{Related Work}
\label{sec:related_work}

\paragraph{Sparse Attention for 
Video Generation Models.}
Sliding Tile Attention~\citep{zhang2025fast},
SVG~\citep{svg}, Sparse-vDiT~\citep{chen2026sparse},
Radial Attention~\citep{li2026radial}, and
LVSA~\citep{glorian2026lvsa} exploit spatial, temporal,
or head-specific attention structure.
SpargeAttention~\citep{zhang2025spargeattention} and
XAttention~\citep{xu2025xattention} estimate block importance online. AdaSpa~\citep{xia2025training} combines online search with cached statistics across denoising steps, while
HEART~\citep{zheng2026haste} reuses sparse masks based on
head-specific changes in queries and keys. ScalingAttention~\citep{zhou2026scalingattention} extracts
sparse topology priors offline. Among trained methods, VORTA~\citep{sun2026vorta}
learns to route between attention variants, VSA~\citep{vsa} trains sparse attention, and
Veda~\citep{han2026veda} uses distillation. VMoBA~\citep{wu2026vmoba} supports sparse training and training-free inference. 
SparsePR combines online token partitioning with an output correction fitted from exact query rows in the current
attention call, addressing shared block selection and error left by skipped interactions.

\paragraph{Token Partitioning for Executable Block Sparsity.}
Earlier methods organize tokens through hashing,
clustering, and learned
permutations~\citep{kitaev2020reformer,roy2021efficient,vyas2020fast,daras2020smyrf,tay2020sparse}. SVG2~\citep{svg2} separately clusters query
and keys into contiguous groups. DraftAttention~\citep{shen2025draftattention} uses
low-resolution attention to guide sparse selection, while
DFSAttn~\citep{hu2026dfsattn} combines Hilbert curve token
reordering with hierarchical block scoring.
AdaCluster~\citep{tan2026adacluster} uses angular similarity for queries and Euclidean similarity for keys.
SVOO~\citep{luo2026attention} alternates key clustering based on query-centroid affinities and query clustering based on key-centroid affinities, using sparsity budgets calibrated offline. Response-Coupled Partitioning derives low-dimensional key coordinates from a metric built from sampled queries, then uses key group centroids to define query coordinates. It requires neither alternating partition refinement nor offline sparsity profiling, operating under a specified executed-pair budget.

\paragraph{Attention Approximation and Residual Reconstruction.}
Attention approximations use random features, landmarks, or sparse and low-rank approximations
\citep{choromanski2020rethinking,xiong2021nystromformer,
chen2021scatterbrain}. 
Re-ttention~\citep{chen2026re} reshapes attention using
past softmax statistics. Rectified
SpaAttn~\citep{liu2025rectified} adjusts attention allocation using pooled attention estimates.
PSA~\citep{li2025psa} represents K/V blocks at multiple
pooling levels. SVG-EAR~\citep{zhou2026svg} combines centroid-based compensation for skipped blocks with error-aware routing. PISA~\citep{li2026pisa} approximates unselected interactions through blockwise Taylor expansions, while Sol-Attn~\citep{li2026sol} uses shared token to block scores for routing and approximate correction. SparsePR fits a call-specific affine map from sparse outputs to the differences between dense and renormalized sparse outputs measured on exact probe rows. Weighted ridge regression fits this map, which corrects unprobed queries directly in output space. This correction models the combined effect of omitted
interactions and softmax renormalization directly in output space.


\section{Sparse Attention Formulation and Observations}
\label{sec:formulation}

We formalize how token partitioning, route selection, and the treatment of skipped interactions define the sparse attention output. We then examine three observations that motivate SparsePR.


\paragraph{Dense and block-sparse attention.}
\label{sec:block_sparse_problem}
For one attention head, let
$Q$, $K$, and $V$
denote the query, key, and value matrices. $Q$ contains $N_q$ query
tokens, while $K$ and $V$ contain $N_k$ paired key and value
tokens, all with per-head dimension $d$. We use row vectors
\(q_i\), \(k_j\), and \(v_j\) for their individual token representations. Dense attention computes
\begin{equation}
A =
\operatorname{softmax}\!\left(
    \frac{QK^\top}{\sqrt{d}}
\right),
\qquad
O^{\mathrm{dense}} = AV .
\label{eq:dense_attention}
\end{equation}

A block-sparse operator partitions the query indices into
$\mathcal{G}^{Q}=\{G_a^Q\}_{a=1}^{C_q}$ and the key indices
into $\mathcal{G}^{K}=\{G_b^K\}_{b=1}^{C_k}$. The value paired with each key inherits its group assignment. The Cartesian
product $G_a^Q\times G_b^K$ defines a cell of query--key
interactions. Let $z_{ab}\in\{0,1\}$ indicate whether this
cell is selected for exact computation. All queries in
$G_a^Q$ share the same selected key groups and their
corresponding values. We define routing density
$\rho_{\mathrm{route}}$ as the number of query and key
interactions evaluated in selected cells divided by
$N_qN_k$. This definition counts individual interactions,
so it accounts for unequal group sizes. Selecting all cells
recovers dense attention and gives
$\rho_{\mathrm{route}}=1$.


Under renormalized hard drop, unselected keys are
omitted and softmax is normalized over the retained keys.
Let $K_a$ and $V_a$ collect the key and value rows from
groups with $z_{ab}=1$. For each query $i\in G_a^Q$,
\begin{equation}
O_i^{\mathrm{sp}}
=
\operatorname{softmax}\!\left(
\frac{q_i K_a^\top}{\sqrt{d}}
\right)V_a .
\label{eq:hard_drop}
\end{equation}
Collecting these rows gives $O^{\mathrm{sp}}$.
We define the residual left by this sparse operator as
\[
R = O^{\mathrm{dense}} - O^{\mathrm{sp}}.
\]

\subsection{Structural Observations on Sparse Attention}
\label{sec:diagnostic_observations}
We use semantic clustering as a reference partition, applying k-means separately to the current query and key activations, with each value token inheriting the assignment of its paired key~\citep{svg2}. We use this partition to make the following three observations:


\par\noindent\textbf{O1: Per-Query Sparsity Does Not Determine Shared-Route Support.}
\label{sec:obs_shared_support}

\begin{figure}[!t]
    \centering
    \includegraphics[width=0.9\textwidth]
    {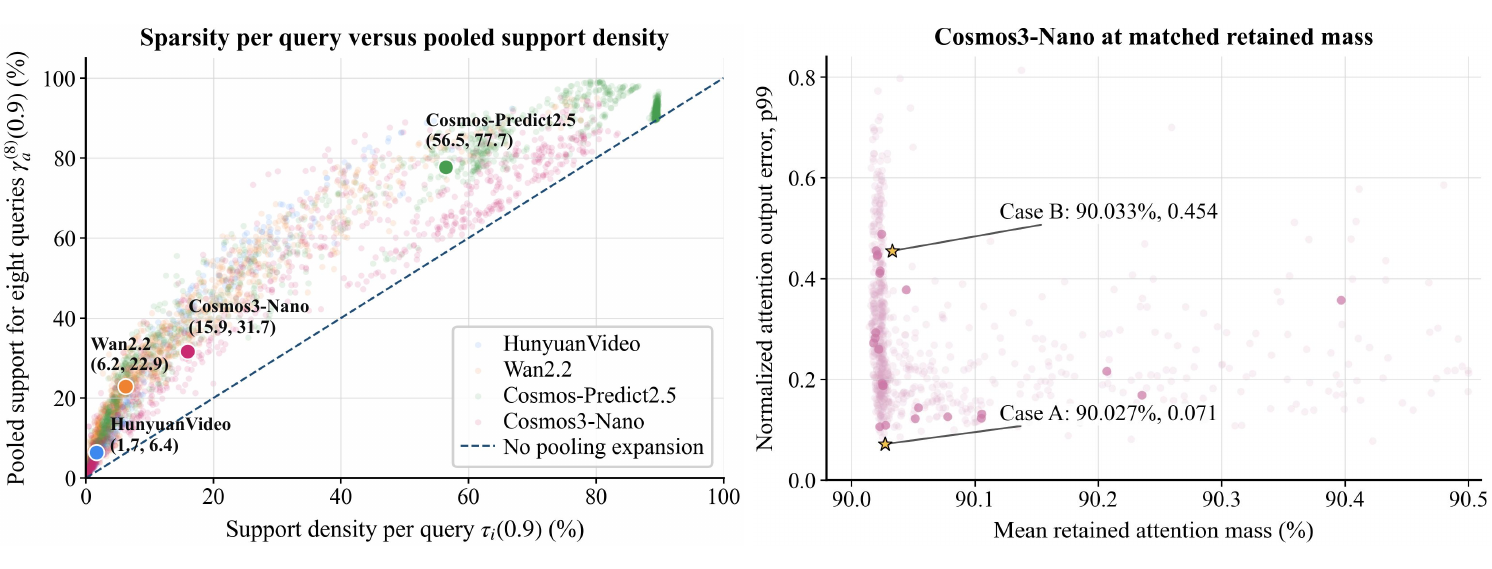}
    \caption{
    Structural observations under the semantic reference partition.
    (a) Pooling eight queries expands the support required to retain
\(90\%\) attention mass. Markers denote model medians.
(b) Nearly identical retained mass can yield substantially different
output errors. 
}
    \label{fig:structural_observations}
\end{figure}

For retained mass \(\alpha\in(0,1]\), let \(D_i(\alpha)\)
denote the smallest set of highest-weight key indices whose cumulative dense attention mass for query \(i\) is at least \(\alpha\). For \(m\) queries \(i_1,\ldots,i_m\in G_a^Q\), define
\[
    \tau_i(\alpha)
    =
    \frac{\lvert D_i(\alpha)\rvert}{N_k},
    \qquad
    \gamma_a^{(m)}(\alpha)
    =
    \frac{
        \left\lvert
        \bigcup_{r=1}^{m}D_{i_r}(\alpha)
        \right\rvert
    }{N_k}.
\]
Here, \(\tau_i(\alpha)\) measures individual support density, while
\(\gamma_a^{(m)}(\alpha)\) measures pooled support density, defined by
the union of individual top-mass supports. This union retains at least
\(\alpha\) attention mass for every sampled query, although a smaller
jointly selected shared support may suffice. By operating on individual
keys, the diagnostic isolates query-side support overlap from the
effects of K/V-block granularity.
Under the semantic partition, Fig.~\ref{fig:structural_observations}(a)
compares both densities at \(\alpha=0.9\). Pooling eight queries
increases the median support density from \(6.2\%\) to \(22.9\%\)
for Wan2.2 and from \(56.5\%\) to \(77.7\%\) for Cosmos-Predict2.5. Pooled support therefore depends on both individual attention concentration and support overlap within query groups.


\par\noindent\textbf{O2: Retained Attention Mass Alone Does Not Determine Output Error.}
\label{sec:obs_output_fidelity}
For \(i\in G_a^Q\), let \(U_a\) be the omitted key indices and
\(p_{U,i}=\sum_{j\in U_a}A_{ij}\) its dense attention mass.
For \(p_{U,i}>0\), let \(O_{U,i}\) be the output formed by
renormalizing the dense attention weights over \(U_a\). The residual then satisfies:
\begin{equation}
R_i
=
O_i^{\mathrm{dense}}-O_i^{\mathrm{sp}}
=
p_{U,i}\left(O_{U,i}-O_i^{\mathrm{sp}}\right).
\label{eq:hard_drop_error_identity}
\end{equation}
The derivation is provided in Appendix~\ref{app:residual_derivation}. Thus, high retained mass makes \(p_{U,i}\) small but does not constrain \(\lVert O_{U,i}-O_i^{\mathrm{sp}}\rVert_2\). Figure~\ref{fig:structural_observations}(b) demonstrates this
for Cosmos3-Nano. Each point summarizes one attention head and reports the mean retained mass and p99 normalized output
error across its queries.
Cases A and B are two heads evaluated for the same input, layer,
and generation step. Their retained masses differ by only
\(0.006\) percentage points, whereas their p99 errors differ by
\(6.4\times\). Appendix~\ref{app:matched_mass_error} reports the
same comparison for the other three models.

\par\noindent\textbf{O3: Partition Choice Affects Affine Residual Representability.}
\label{sec:obs_calibration} We examine whether changing the partition affects how well an affine function of the sparse output can represent the attention output error. Let \(g\) index a query and K/V partition configuration, with sparse output \(O_g^{\mathrm{sp}}\) and residual
\(R_g=O^{\mathrm{dense}}-O_g^{\mathrm{sp}}\).
We fit an affine estimate:
\begin{equation}
    \widehat R_g^{\mathrm{oracle}}
    = O_g^{\mathrm{sp}}B_g+\mathbf{1}b_g.
\end{equation}
Here, \(B_g\) maps sparse output channels to residual channels, and the row vector \(b_g\) is an offset shared across queries. We choose the coefficients to minimize the
squared fitting error summed over all evaluated queries and output channels. We call this an oracle fit because it uses dense outputs on all evaluated queries, rather than only a small probe set.

Across four evaluated models, response-coupled partitioning increases the fraction of residual squared norm explained by the best affine fit by 3.70--15.52 percentage points and reduces the remaining squared error by 36.6--70.8\% relative to semantic
partitioning. Thus, the fit captures a larger fraction of the residual and leaves less error. The smaller remaining error reflects both the initial residual
magnitude and the fraction explained by the fit. Appendix~\ref{app:oracle_affine_representability} reports
results for each model. Section~\ref{sec:residual_repair} describes estimating a
correction within the same affine family from a small probe set using weighted ridge regression, with accuracy evaluated separately on unprobed queries.
\section{Method}
\label{sec:method}

Following observations in Section~\ref{sec:formulation}, SparsePR combines
token grouping based on responses between queries and keys
with an affine correction fitted from exact probe rows. Figure~\ref{fig:main} summarizes the two components of SparsePR.



\subsection{Response-Coupled Partitioning}
\label{sec:r_clustering}

An executable partition should group queries with similar routing behavior and paired K/V tokens whose keys induce similar response profiles. Proximity in the activation space does not guarantee either property. Nearby queries can require different K/V supports, while nearby keys can respond differently under the same queries. We construct partitions in low-dimensional response coordinates
derived from the current attention call, capturing the dominant
response variations while reducing the cost of clustering. The construction is coupled and asymmetric. Sampled queries define the coordinates for K/V grouping, and the resulting groups define the query coordinates.


\par\noindent\textbf{Key-response coordinates for paired K/V groups.}

Let \(Q_s\in\mathbb{R}^{M_s\times d}\) contain \(M_s\) query rows selected deterministically at evenly spaced token positions
from the current attention call. We group keys by the similarity
of their responses to these queries using the metric
\[
    M_K
    =
    \frac{1}{M_s}Q_s^\top Q_s .
\]
For any two keys \(k_j\) and \(k_\ell\),
\[
(k_j-k_\ell)M_K(k_j-k_\ell)^\top
=
\frac{1}{M_s}
\left\|Q_s(k_j-k_\ell)^\top\right\|_2^2.
\]
Thus, distance under \(M_K\) measures the average squared
response difference between keys: keys are close when they
respond similarly across the sampled queries.
These responses are pre-softmax logits up to the common
\(1/\sqrt d\) attention scale.
Appendix~\ref{app:key_response_coordinates} provides the derivation.

To obtain compact clustering coordinates, we retain the leading
\(r_K\) eigenvectors of \(M_K\). We normalize their eigenvalues
to unit mean and weight each eigenvector by the square root
of its normalized eigenvalue, forming
\(F_K\in\mathbb{R}^{d\times r_K}\).
We represent key \(k_j\) by
\begin{equation}
    \phi_K(k_j)
    =
    k_jF_K
    \in\mathbb{R}^{r_K}.
    \label{eq:key_response_feature}
\end{equation}

Squared Euclidean distances in these coordinates reproduce
the rank-\(r_K\) truncated response metric up to a common
positive scale. Applying \(k\)-means with \(C_k\) requested
clusters yields the paired K/V groups \(\mathcal{G}^{KV}\),
with each value inheriting its key's assignment. In the centroid-based
construction below, \(C_k\) denotes the number of nonempty K/V groups. 

\par\noindent\textbf{Query-response coordinates from K/V groups.}
The resulting K/V groups define the response space used to
compare queries. For each nonempty group, we compute its centroid
in the original key space. We subtract the uniform mean of
these centroids and stack the centered vectors as the rows of \(\widetilde K\).
Centering removes a common offset from each query's logits
across groups while preserving their relative differences.
The query-response metric is
\[
    M_Q=\frac{1}{C_kd}\widetilde K^\top\widetilde K,
\]
Two queries are close under this metric when they produce
similar relative pre-softmax logit profiles over these groups. Applying the same eigenvalue weighted construction to the
leading \(r_Q\) eigendirections of \(M_Q\) gives
\(F_Q\in\mathbb{R}^{d\times r_Q}\).
Each projected query is then RMS-normalized:
\begin{equation}
    \phi_Q(q_i)
    =
    \operatorname{RMSNorm}(q_iF_Q).
    \label{eq:query_response_feature}
\end{equation}
RMS normalization divides each projected vector by its root
mean square magnitude, emphasizing response direction and
reducing the influence of magnitude. Numerical stabilization
is specified in Appendix~\ref{app:query_response_coordinates}.
The original queries are retained
for routing and attention. Applying \(k\)-means with \(C_q\) requested clusters to
\(\phi_Q(q_i)\) yields the query partition \(\mathcal G^Q\)
used for shared routing. The construction proceeds once
from K/V grouping to query grouping, without alternating
refinement between the two partitions.

\subsection{Probe-Fitted Residual Reconstruction}
\label{sec:residual_repair}

For each attention call, Probe-Fitted Residual Reconstruction uses
\(M \ll N_q\) exact query rows to fit a residual estimator.
Motivated by the affine diagnostic in Section~\ref{sec:obs_calibration},
we model the residual as an affine function of the sparse output:
\[
x_i = O_i^{\mathrm{sp}} \in \mathbb{R}^{d},
\qquad
R_i \approx b + x_i B.
\]

\paragraph{Exact probe calibration.}
We select \(M\) probe indices
\(\mathcal P\subseteq\{1,\ldots,N_q\}\) using the query groups.
Exact attention on each probe gives
\(R_p=O_p^{\mathrm{dense}}-O_p^{\mathrm{sp}}\).
Let \(X_{\mathcal P},R_{\mathcal P}\in\mathbb{R}^{M\times d}\)
stack the probe sparse outputs and residuals, respectively,
and let \(W\) be the diagonal matrix of group-coverage weights
normalized to unit mean. These weights reflect the populations
of the represented query groups.
Groups without probes contribute no residual observations
to the fit, but their queries use the same fitted correction.
The selection and coverage-weight rules are specified in
Appendix~\ref{app:probe_selection}; weight normalization is given in
Appendix~\ref{app:regularized_fit}.
Using weighted probe statistics, we standardize
\(X_{\mathcal P}\) and center \(R_{\mathcal P}\), yielding
\(\bar X_{\mathcal P}\) and \(\bar R_{\mathcal P}\).
We denote the weighted residual mean by \(\mu_R\).

\par\noindent\textbf{Weighted ridge fit.}
We estimate the feature-to-residual map by weighted ridge
regression:
\begin{equation}
    B_\lambda
    =
    \arg\min_B
    \left\|
        W^{1/2}
        \left(
            \bar R_{\mathcal P}
            -
            \bar X_{\mathcal P}B
        \right)
    \right\|_F^2
    +
    \lambda\|B\|_F^2 .
    \label{eq:probe_ridge_fit}
\end{equation}
Here, \(B_\lambda\in\mathbb{R}^{d\times d}\) maps
standardized sparse-output features to centered residual
channels, and \(\lambda>0\) controls the regularization strength.
The weights determine each probe's contribution to the fit,
while the ridge penalty shrinks the coefficients and
stabilizes estimation from a small probe set.
Standardization and the solution procedure are detailed in
Appendix~\ref{app:regularized_fit}.


\par\noindent\textbf{Residual prediction.}
For an unprobed query, let \(\bar x_i\) denote its sparse
output standardized using the probe statistics. We predict
\(\widehat R_i=\mu_R+\bar x_iB_\lambda\), without output-subspace
truncation (Appendix~\ref{app:probe_sensitivity}).
The final attention output is

\begin{equation}
    \widehat O_i
    =
    \begin{cases}
        O_i^{\mathrm{dense}},
        & i\in\mathcal P,\\[1mm]
        O_i^{\mathrm{sp}}+\widehat R_i,
        & i\notin\mathcal P.
    \end{cases}
    \label{eq:final_reconstructed_output}
\end{equation}





\subsection{Sparse Execution and Exact-Pair Accounting}
\label{sec:kernel_budget}

\paragraph{Routing and GQA.} 
After reserving the cost of exact probes, we convert the target
routed density into a common per-query key-token budget.
For each query group, we rank nonempty K/V groups in descending
order of the value-error score defined in Appendix~\ref{app:routing_budget}
and retain the shortest ranked prefix whose cumulative key
count reaches this budget. Each selected cell contributes
\(\lvert G_a^Q\rvert\lvert G_b^{KV}\rvert\) query--key pairs.
Whole-cell selection can cause realized density to differ
from the target, so we report it for each configuration. For grouped-query attention (GQA), the key-response metric
and K/V partition are shared across query heads associated
with the same native K/V head. Query partitions and routing
decisions are computed separately for each query head.
\par\noindent\textbf{Sparse execution and pair accounting.} We permute \(Q\), \(K\), and \(V\) into contiguous group-major
layouts, keeping each value token with its paired key.
FlashInfer's variable-block sparse-attention
primitive~\citep{ye2025flashinfer} takes the selected cells
and ragged group offsets, evaluates the selected interactions
exactly, and renormalizes each query row over its retained
K/V groups as in Eq.~(\ref{eq:hard_drop}).
We then restore the output to the original query order. 
After sparse attention, we evaluate \(M\) exact probe rows
per head against all keys, giving
\(\rho_{\mathrm{exec}}=\rho_{\mathrm{route}}+M/N_q\),
where \(N_q\) is the query count per head. This counts
pairs recomputed during probe evaluation.




\section{Experiments}
\label{sec:experiments}

\noindent\textbf{Implementation details.}
We evaluate all models on a single NVIDIA H100 with BF16
activations. SparsePR uses \(C_q=300\), \(C_k=1000\),
\(r_K=48\), \(r_Q=64\), \(M=64\), and \(\lambda=1\).
The total executed-pair target, including probes, is 22\%
for HunyuanVideo, Wan2.2, and Cosmos-Predict2.5, and 26\%
for Cosmos3-Nano. Timings include all online overheads.
Appendix~\ref{app:implementation_protocol} details numerical
precision, query sampling, clustering, GQA, and timing.

\par\noindent\textbf{Models and benchmarks.}
We evaluate HunyuanVideo-13B for text-to-video generation
using prompts from the Penguin suite and VBench metrics.
We evaluate Wan2.2-I2V-A14B for image-to-video generation
using VBench-I2V, the image-to-video component of
VBench++~\cite{huang2025vbench++}.
For Cosmos-Predict2.5-14B and Cosmos3-Nano-16B, we report
selected VBench-I2V metrics and the PBench Quality
score~\cite{nvidia2025pbench} on their respective
evaluation subsets.


\par\noindent\textbf{Metrics.}
We measure fidelity to dense outputs using PSNR, SSIM,
and LPIPS on corresponding frames. Generation quality
is measured by VBench/VBench-I2V imaging quality (\emph{ImgQual}),
subject consistency, and PBench Quality.
We report realized executed-pair density, end-to-end
FLOPs, and end-to-end generation speedup.
Dense, sparse, and reproduced baseline runs use matched
inputs, seeds, and generation settings.

\par\noindent\textbf{Quality and efficiency.}
Table~\ref{tab:main_results} compares dense-reference fidelity,
generation quality, and efficiency across four models.
Figure~\ref{fig:qualitative} shows matched dense and SparsePR
outputs, with additional comparisons in
Appendix~\ref{app:qualitative}.

\begin{table*}[t]
\centering
\caption{
Quality--efficiency comparison across content-generation and physical
video world models. 
}
\label{tab:main_results}

\setlength{\tabcolsep}{4.2pt}
\renewcommand{\arraystretch}{0.96}

\resizebox{\textwidth}{!}{%
\begin{tabular}{@{}ll ccc cc c ccc@{}}
\toprule
& &
\multicolumn{3}{c}{\textbf{Dense-Reference Fidelity}} &
\multicolumn{2}{c}{\textbf{VBench / VBench++}} &
\multicolumn{1}{c}{\textbf{PBench}} &
\multicolumn{3}{c}{\textbf{Efficiency}} \\
\cmidrule(lr){3-5}
\cmidrule(lr){6-7}
\cmidrule(lr){8-8}
\cmidrule(lr){9-11}

\textbf{Model} &
\textbf{Method} &
\textbf{PSNR$\uparrow$} &
\textbf{SSIM$\uparrow$} &
\textbf{LPIPS$\downarrow$} &
\textbf{ImgQual$\uparrow$} &
\textbf{SubCons$\uparrow$} &
\textbf{Quality$\uparrow$} &
\textbf{Density$\downarrow$} &
\textbf{PFLOPs$\downarrow$} &
\textbf{E2E Speedup$\uparrow$} \\
\midrule


\multicolumn{11}{@{}l}{\textit{Content-generation video models}} \\
\addlinespace[1pt]


\multirow{6}{*}{
\makecell[l]{
\textit{HunyuanVideo}\\
13B, 720p\\
Text-to-Video
}}
& Dense
& --
& --
& --
& 0.850
& 0.976
& --
& 100.0\%
& 612.38
& 1.00$\times$ \\

& SpargeAttn~\cite{zhang2025spargeattention}
& 25.31
& 0.832
& 0.217
& 0.763
& \underline{0.928}
& --
& 40.19\%
& 389.76
& 1.38$\times$ \\

& SVG2~\cite{svg2}
& 29.87
& 0.907
& 0.121
& \textbf{0.850}
& 0.927
& --
& 25.45\%
& 299.02
& \underline{2.30$\times$} \\

& SVOO~\cite{luo2026attention}
& 24.87
& 0.843
& 0.224
& 0.679
& \textbf{0.976}
& --
& 33.26\%
& 345.81
& 2.17$\times$ \\

& SVG-EAR~\cite{zhou2026svg}
& \underline{30.54}
& \underline{0.918}
& \underline{0.098}
& \underline{0.845}
& 0.903
& --
& \underline{22.17\%}
& \underline{281.86}
& 1.93$\times$ \\

\rowcolor{NVIDIALight}
& \textbf{SparsePR}
& \textbf{31.84}
& \textbf{0.932}
& \textbf{0.087}
& \textbf{0.850}
& \textbf{0.976}
& --
& \textbf{21.92\%}
& \textbf{255.95}
& \textbf{2.61$\times$} \\

\midrule


\multirow{6}{*}{
\makecell[l]{
\textit{Wan2.2-I2V-A14B}\\
14B active, 720p\\
Image-to-Video
}}
& Dense
& --
& --
& --
& 0.689
& 0.974
& --
& 100.0\%
& 658.46
& 1.00$\times$ \\

& SpargeAttn~\cite{zhang2025spargeattention}
& 26.74
& 0.871
& 0.116
& 0.680
& 0.953
& --
& 30.15\%
& 396.83
& 1.58$\times$ \\

& SVG2~\cite{svg2}
& 28.18
& 0.878
& 0.105
& 0.668
& \underline{0.970}
& --
& 31.28\%
& 393.95
& 1.59$\times$ \\

& SVOO~\cite{luo2026attention}
& 29.67
& \underline{0.913}
& 0.095
& \textbf{0.689}
& \textbf{0.973}
& --
& 31.67\%
& 389.62
& \underline{1.61$\times$} \\

& SVG-EAR~\cite{zhou2026svg}
& \underline{29.75}
& \textbf{0.921}
& \underline{0.086}
& 0.680
& \underline{0.970}
& --
& \underline{23.64\%}
& \underline{378.88}
& \underline{1.61$\times$} \\

\rowcolor{NVIDIALight}
& \textbf{SparsePR}
& \textbf{30.66}
& 0.907
& \textbf{0.044}
& \underline{0.687}
& \textbf{0.973}
& --
& \textbf{21.97\%}
& \textbf{336.55}
& \textbf{1.80$\times$} \\

\midrule


\multicolumn{11}{@{}l}{\textit{Physical video world models}} \\
\addlinespace[1pt]


\multirow{4}{*}{
\makecell[l]{
\textit{Cosmos-Predict2.5}\\
14B, 720p\\
\textnormal{Image-to-World}
}}
& Dense
& --
& --
& --
& 0.714
& 0.976
& 77.76
& 100.0\%
& 526.87
& 1.00$\times$ \\

& SVG2~\cite{svg2}
& 20.07
& 0.624
& 0.330
& 0.678 
& 0.896
& 76.14 
& \underline{28.81\%} 
& \underline{286.51}
& \underline{1.24$\times$}  \\

& SVOO~\cite{luo2026attention}
& 22.06
& 0.685
& 0.289
& 0.701
& \underline{0.909}
& 76.03
& 37.63\% 
& 315.38
& 1.03$\times$ \\

& SVG-EAR~\cite{zhou2026svg}
& \underline{25.54}
& \underline{0.908}
& \textbf{0.062}
& \underline{0.710}
& \textbf{0.976} 
& \textbf{77.76}
& 29.75\% 
& 289.69
& 1.10$\times$ \\

\rowcolor{NVIDIALight}
& \textbf{SparsePR}
& \textbf{26.33}
& \textbf{0.942}
& \underline{0.068}
& \textbf{0.714}
& \textbf{0.976}
& \underline{77.75}
& \textbf{22.14\%}
& \textbf{253.61}
& \textbf{1.51$\times$} \\

\midrule


\multirow{4}{*}{
\makecell[l]{
\textit{Cosmos3-Nano}\\
16B, 720p\\
\textnormal{Image-to-World}
}}
& Dense
& --
& --
& --
& 0.700
& 0.950
& 77.31
& 100.0\%
& 127.61
& 1.00$\times$ \\


& SVG2~\cite{svg2}
& \underline{22.45}
& \underline{0.735}
& \underline{0.216}
& 0.677
& 0.915
& 75.03
& 37.29\%
& 96.19
& \underline{1.16$\times$} \\

& SVOO~\cite{luo2026attention}
& 16.64
& 0.573
& 0.381
& \textbf{0.707}
& \textbf{0.962}
& \textbf{77.59}
& 67.32\%
& 108.01
& 1.02$\times$ \\

& SVG-EAR~\cite{zhou2026svg}
& 21.16
& 0.709
& 0.261
& 0.658
& 0.872
& 72.85
& \underline{37.18\%}
& \underline{96.14}
& 1.10$\times$ \\

\rowcolor{NVIDIALight}
& \textbf{SparsePR}
& \textbf{24.42}
& \textbf{0.801}
& \textbf{0.176}
& \underline{0.699}
& \underline{0.949}
& \underline{77.30}
& \textbf{25.96\%}
& \textbf{81.79}
& \textbf{1.48$\times$} \\
\bottomrule
\end{tabular}%
}
\vspace{2pt}
\parbox{\textwidth}{
}
\vspace{-4pt}
\end{table*}


\subsection{Ablations}
\label{sec:ablations}

\noindent\textbf{Partition and reconstruction.}
Table~\ref{tab:partition_repair_ablation} compares partitioning and residual reconstruction.

\begin{table}[!htb]
\centering
\caption{
Partition and reconstruction ablations at 22\% executed-pair density, including exact probes. Entries report mean/p99 normalized attention output error on unprobed rows. Lower is better. Repair comparisons fix probe indices and estimator settings. The key-response K/V variant changes only K/V grouping, retaining semantic query groups.
}
\label{tab:partition_repair_ablation}
{\small
\setlength{\tabcolsep}{4pt}
\renewcommand{\arraystretch}{0.96}
\begin{tabular}{@{}lcccc@{}}
\toprule
\textbf{Configuration}
& \makecell{\textbf{HunyuanVideo}}
& \makecell{\textbf{Wan2.2}}
& \makecell{\textbf{Cosmos-}\\\textbf{Predict2.5}}
& \makecell{\textbf{Cosmos3-Nano}}
\\
\midrule

\multicolumn{5}{@{}l}{\textit{Partition construction, repair disabled}} \\
\addlinespace[1pt]

Semantic partition
& 0.089 / \underline{0.714}
& 0.164 / 1.734
& 0.791 / 7.532
& 0.359 / 3.356
\\
Key-response K/V partition
& \underline{0.085} / 0.812
& \underline{0.156} / \underline{1.659}
& \underline{0.769} / \underline{7.270}
& \underline{0.341} / \underline{3.174}
\\

\textbf{Response-coupled partition}
& \textbf{0.073 / 0.696}
& \textbf{0.108 / 0.647}
& \textbf{0.761 / 7.227}
& \textbf{0.331 / 3.150}
\\
\midrule
\multicolumn{5}{@{}l}{\textit{Partition--repair interaction}} \\
\addlinespace[1pt]
Semantic partition + probe repair
& \underline{0.053} / \underline{0.356}
& \underline{0.105} / \underline{0.819}
& \underline{0.262} / \underline{0.826}
& \underline{0.124} / \underline{0.668}
\\
\textbf{SparsePR}
& \textbf{0.030 / 0.207}
& \textbf{0.063 / 0.323}
& \textbf{0.075 / 0.489}
& \textbf{0.076 / 0.447}
\\
\bottomrule
\end{tabular}
}
\end{table}

With repair disabled, response-coupled partitioning lowers
mean and p99 error relative to semantic partitioning across
all four models. Both components reduce error, and their combination achieves
the lowest mean and p99 error across all four models. With the same probes and estimator,
SparsePR reduces mean error by 38.7--71.4\% and p99 error
by 33.1--60.6\% relative to semantic partitioning with repair
(Fig.~\ref{fig:partition_runtime}(a)).


\par\noindent\textbf{Density sweep.}
Table~\ref{tab:cosmos3_density_sweep} varies the total executed-pair budget for Cosmos3-Nano from 12\% to 35\%, holding remaining settings fixed within each method. Exact probes count toward the repaired methods' budgets. SparsePR achieves the lowest values among the compared methods at all five budgets.

\begin{table}[!htb]
\centering
\caption{
Cosmos3-Nano density sweep. Within each method, only the
executed-pair budget varies, including exact probes where repair is used. Entries report mean/p99 normalized
attention-output error on unprobed rows. Lower is better.
}
\label{tab:cosmos3_density_sweep}

{\footnotesize
\setlength{\tabcolsep}{2.2pt}
\renewcommand{\arraystretch}{0.98}

\begin{adjustbox}{max width=\textwidth,center}
\begin{tabular}{@{}lccccc@{}}
\toprule
\textbf{Method}
& \(\boldsymbol{\rho=12\%}\)
& \(\boldsymbol{\rho=17\%}\)
& \(\boldsymbol{\rho=22\%}\)
& \(\boldsymbol{\rho=28\%}\)
& \(\boldsymbol{\rho=35\%}\)
\\
\midrule

Semantic + hard drop
& 0.547 / 4.993
& 0.435 / 4.010
& 0.359 / 3.356
& 0.293 / 2.778
& 0.235 / 2.266
\\

Response-coupled + hard drop
& 0.506 / 4.634
& 0.402 / 3.749
& 0.331 / 3.150
& 0.270 / 2.616
& 0.217 / 2.142
\\

Semantic + probe repair
& \underline{0.172} / \underline{0.865}
& \underline{0.148} / \underline{0.838}
& \underline{0.124} / \underline{0.668}
& \underline{0.076} / \underline{0.484}
& \underline{0.063} / \underline{0.418}
\\
\textbf{SparsePR}
& \textbf{0.127 / 0.724}
& \textbf{0.099 / 0.577}
& \textbf{0.076 / 0.447}
& \textbf{0.068 / 0.422}
& \textbf{0.056 / 0.366}
\\

\bottomrule
\end{tabular}
\end{adjustbox}
}
\end{table}


\par\noindent\textbf{Probe and estimator design.}
In Table~\ref{tab:probe_estimator_ablation}, query-group-stratified
probes achieve the lowest mean and p99 errors across four models,
tying random probes in mean error on Cosmos-Predict2.5.
With sparse outputs, probes, and evaluation rows fixed,
full weighted ridge outperforms mean-only, scalar affine,
and diagonal affine corrections in both metrics, supporting
cross-channel residual prediction.
Further appendices~\ref{app:query_rms_ablation}
and~\ref{app:probe_sensitivity} examine query RMS normalization
and sensitivity to probe count, output-rank truncation,
and ridge strength.

\begin{table}[!htb]
\centering
\caption{
Probe-selection and correction-family ablations with response-coupled
partitioning, $M=64$, $\lambda=1$.
Within each panel, only the indicated component varies.
Entries report mean/p99 normalized attention-output error on a
shared set of unprobed query rows. Lower is better. Bold row labels mark the reference
configuration, repeated in both panels.
}
\label{tab:probe_estimator_ablation}

{\small
\setlength{\tabcolsep}{4pt}
\renewcommand{\arraystretch}{0.96}
\begin{tabular}{@{}lcccc@{}}
\toprule
\textbf{Configuration}
& \makecell{\textbf{HunyuanVideo}}
& \makecell{\textbf{Wan2.2}}
& \makecell{\textbf{Cosmos-}\\\textbf{Predict2.5}}
& \makecell{\textbf{Cosmos3-Nano}}
\\
\midrule
\multicolumn{5}{@{}l}{\textit{Probe selection: full weighted ridge fixed}} \\
\addlinespace[1pt]
Random probes & 0.039 / \underline{0.235} & \underline{0.080} / \underline{0.356} & \textbf{0.075} / \underline{0.503} & \underline{0.083} / \underline{0.467} \\
Uniform spatiotemporal probes & \underline{0.037} / 0.246 & 0.089 / 0.394 & \underline{0.090} / 0.543 & 0.096 / 0.549 \\
\textbf{Query group stratified probes} & \textbf{0.030 / 0.207} & \textbf{0.063 / 0.323} & \textbf{0.075 / 0.489} & \textbf{0.076 / 0.447} \\
\midrule
\multicolumn{5}{@{}l}{\textit{Correction family: query group stratified probes fixed}} \\
\addlinespace[1pt]
Mean-only & 0.043 / 0.362 & 0.089 / 0.455 & 0.350 / 3.538 & 0.191 / 1.376 \\
Scalar affine & 0.034 / 0.281 & 0.068 / 0.365 & 0.195 / 2.069 & 0.131 / 0.880 \\
Diagonal affine & \underline{0.032} / \underline{0.246} & \underline{0.065} / \underline{0.352} & \underline{0.082} / \underline{0.543} & \underline{0.099} / \underline{0.604} \\
\textbf{Full weighted ridge} & \textbf{0.030 / 0.207} & \textbf{0.063 / 0.323} & \textbf{0.075 / 0.489} & \textbf{0.076 / 0.447} \\
\bottomrule
\end{tabular}
}
\end{table}





\subsection{Runtime Analysis}
\label{sec:runtime}

\begin{figure}[!t]
    \centering
    \vspace{6pt}
    \includegraphics[width=0.9\textwidth]{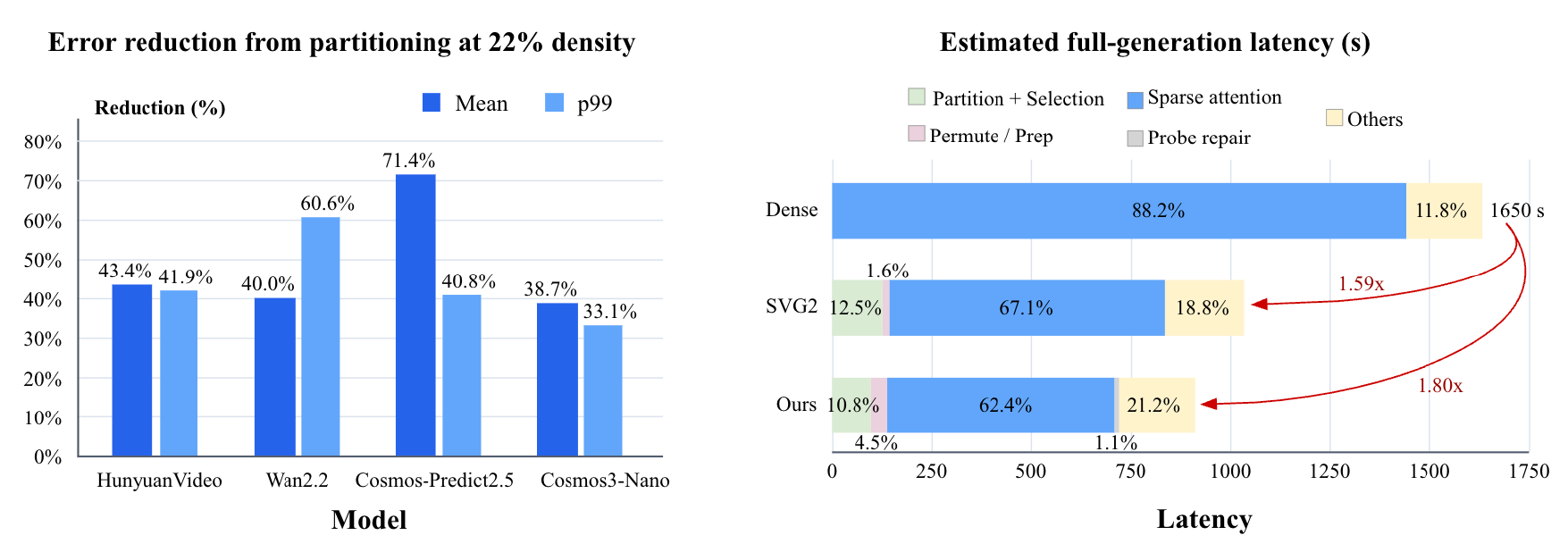}
    \caption{
Partitioning and runtime analysis. (a) Error reduction from response-coupled vs. semantic partitioning at 22\% density. (b) Wan2.2 full generation latency for Dense, SVG2, and SparsePR.
}
    \label{fig:partition_runtime}
    \vspace{-6pt}
\end{figure}

Figure~\ref{fig:partition_runtime}(b) decomposes estimated
full-generation latency on Wan2.2. SparsePR reduces latency from \(1650\)~s to \(917\)~s, corresponding to a \(1.80\times\) speedup, compared with \(1038\)~s and \(1.59\times\) for SVG2. Probe-Fitted Residual
Reconstruction accounts for only \(1.1\%\) of latency. Thus, the primary fidelity gain from probe fitting adds little end-to-end overhead.

\FloatBarrier
\section{Conclusion}
\label{sec:conclusion}
We study training-free block-sparse attention as an executable
operator in which partitioning shapes both shared support and the
output residual left by skipped interactions. SparsePR uses
Response-Coupled Partitioning to improve support overlap under shared
routing and Probe-Fitted Residual Reconstruction to correct the output
residual left by skipped interactions. Across four video generation
and world models, SparsePR reduces mean attention-output error on
unprobed queries by 61.6--90.5\% relative to semantic block-sparse
attention without correction at 22\% executed-pair density, including
probes. Ablations show that response-coupled partitioning reduces
error before correction and improves reconstruction under the same
probe budget. SparsePR achieves \(1.48\times\)--\(2.61\times\)
end-to-end speedups while maintaining benchmark quality close to
dense attention at 21.9--26.0\% realized executed-pair density.
These results show that effective sparse execution depends on both
partition geometry and correction of the output error caused by
skipped interactions.


\bibliography{iclr2027_conference}
\bibliographystyle{iclr2027_conference}

\appendix
\clearpage
\addtocontents{toc}{\protect\setcounter{tocdepth}{2}}
\begingroup
\renewcommand{\contentsname}{Table of Contents}
\hypersetup{hidelinks,linktoc=section}
\let\appendixcontentsline\contentsline
\renewcommand{\contentsline}[4]{%
  \appendixcontentsline{#1}{\textcolor{blue!50!black}{#2}}{#3}{#4}}
\tableofcontents
\endgroup
\clearpage
\appendix
\numberwithin{equation}{section}
\numberwithin{table}{section}
\numberwithin{figure}{section}
\renewcommand{\thealgorithm}{\thesection.\arabic{algorithm}}

\section{Residual Reconstruction Details}
\label{app:residual_details}

We derive the residual left by renormalized hard drop and describe the
estimator fitted for each attention call. The estimator uses only sparse
outputs as regression features. Query response coordinates determine
query groups and probe selection but do not enter the regression.

\subsection{Hard Drop Residual and Affine Model}
\label{app:residual_derivation}

For query row $i$ let $\mathcal S_i$ and $\mathcal U_i$ denote the retained
and omitted key sets. The attention logit is
\begin{equation}
    \ell_{ij}=\frac{q_i k_j^\top}{\sqrt d}.
    \label{eq:app_attention_logit}
\end{equation}
The retained and omitted partition sums and value numerators are
\begin{align}
    Z_{S,i}&=\sum_{j\in\mathcal S_i}\exp(\ell_{ij}),
    &
    N_{S,i}&=\sum_{j\in\mathcal S_i}\exp(\ell_{ij})v_j,
    \nonumber\\
    Z_{U,i}&=\sum_{j\in\mathcal U_i}\exp(\ell_{ij}),
    &
    N_{U,i}&=\sum_{j\in\mathcal U_i}\exp(\ell_{ij})v_j.
    \label{eq:app_partition_statistics}
\end{align}
Dense attention and renormalized hard drop therefore compute
\begin{equation}
    O_i^{\mathrm{dense}}
    =
    \frac{N_{S,i}+N_{U,i}}{Z_{S,i}+Z_{U,i}},
    \qquad
    O_i^{\mathrm{sp}}
    =
    \frac{N_{S,i}}{Z_{S,i}}.
    \label{eq:app_dense_sparse_outputs}
\end{equation}
Let
\begin{equation}
    p_{U,i}
    =
    \frac{Z_{U,i}}{Z_{S,i}+Z_{U,i}},
    \qquad
    O_{U,i}
    =
    \frac{N_{U,i}}{Z_{U,i}}
    \quad (p_{U,i}>0)
    \label{eq:app_omitted_statistics}
\end{equation}
denote the omitted mass under dense softmax and the output obtained by
renormalizing over the omitted keys. Then
\begin{equation}
    O_i^{\mathrm{dense}}
    =
    (1-p_{U,i})O_i^{\mathrm{sp}}+p_{U,i}O_{U,i},
\end{equation}
and the exact attention output residual is
\begin{equation}
    \boxed{
    R_i
    =
    O_i^{\mathrm{dense}}-O_i^{\mathrm{sp}}
    =
    p_{U,i}\left(O_{U,i}-O_i^{\mathrm{sp}}\right).
    }
    \label{eq:app_exact_hard_drop_residual}
\end{equation}
Sparse execution provides $O_i^{\mathrm{sp}}$ but does not evaluate the
omitted interactions needed for $p_{U,i}$ and $O_{U,i}$. We fit an affine
map from sparse outputs to measured residuals within the current call.
\begin{equation}
    x_i=O_i^{\mathrm{sp}}\in\mathbb R^{d},
    \qquad
    R_i\approx b+x_iB.
    \label{eq:app_local_affine_model}
\end{equation}
This empirical model does not assume that $O_i^{\mathrm{sp}}$ uniquely
determines the omitted statistics. The oracle analysis measures how much
of the residual this affine family can represent under each partition.

\subsection{Exact Probe Selection Stratified by Query Group}
\label{app:probe_selection}

We fit the affine map for each query head in the current attention call.
The fit therefore adapts to the input and routing pattern at that layer
and denoising step without an offline calibration dataset. We evaluate
$M=64\ll N_q$ probe rows against all keys. Probes add $M/N_q$ to the
executed pair density. This fraction ranges from $0.054\%$ to $0.145\%$
for the evaluated sequence lengths.

Probe selection uses the query groups and their response coordinates
without inspecting dense residuals. For each nonempty group $G_a^Q$ define
\begin{equation}
    c_a
    =
    \frac{1}{|G_a^Q|}
    \sum_{i\in G_a^Q}\phi_Q(q_i),
    \qquad
    d_i
    =
    \left\|\phi_Q(q_i)-c_a\right\|_2.
    \label{eq:app_query_group_distance}
\end{equation}
We order rows within each group by increasing $d_i$ and break ties by
original row index. We order nonempty groups by decreasing size and
break ties by group index. Each traversal takes the next row from every
group in this order and skips exhausted groups. Selection stops after
$M$ distinct probes. With at least $M$ nonempty groups this selects the
row nearest the centroid from each of the $M$ largest groups.

\begin{algorithm}[H]
\caption{Exact Probe Selection Stratified by Query Group}
\label{alg:app_probe_selection}
\small
\begin{algorithmic}[1]
\Require Query partition $\mathcal G^Q$ of $\{1,\ldots,N_q\}$ with
coordinates $\{\phi_Q(q_i)\}_{i=1}^{N_q}$ and budget $1\leq M\leq N_q$
\Ensure $M$ distinct probe indices $\mathcal P$ and diagonal weights $W$ with unit mean
\State $\mathcal A\gets\{a:|G_a^Q|>0\}$ ordered by decreasing $|G_a^Q|$ then increasing $a$
\ForAll{$a\in\mathcal A$}
    \State $c_a\gets |G_a^Q|^{-1}\sum_{i\in G_a^Q}\phi_Q(q_i)$
    \State $d_i\gets\lVert\phi_Q(q_i)-c_a\rVert_2$ for each $i\in G_a^Q$
    \State $\pi_a\gets$ indices in $G_a^Q$ sorted lexicographically by $(d_i,i)$
\EndFor
\State $\mathcal P\gets\varnothing$ and $t\gets 1$
\While{$|\mathcal P|<M$}
    \ForAll{$a\in\mathcal A$ in the specified order}
        \If{$t\leq |G_a^Q|$}
            \State $\mathcal P\gets\mathcal P\cup\{\pi_a(t)\}$
            \If{$|\mathcal P|=M$}
                \State \textbf{break}
            \EndIf
        \EndIf
    \EndFor
    \State $t\gets t+1$
\EndWhile
\ForAll{$a\in\mathcal A$ with $m_a=|\mathcal P\cap G_a^Q|>0$}
    \State $w_p\gets |G_a^Q|/m_a$ for each $p\in\mathcal P\cap G_a^Q$
\EndFor
\State $\bar w\gets M^{-1}\sum_{p\in\mathcal P}w_p$
\State $W\gets\operatorname{diag}(w_p/\bar w)_{p\in\mathcal P}$ in the probe matrix row order
\State \Return $\mathcal P,W$
\end{algorithmic}
\end{algorithm}

Exact attention on each probe $p\in\mathcal P$ provides
\begin{equation}
    R_p
    =
    O_p^{\mathrm{dense}}-O_p^{\mathrm{sp}}.
    \label{eq:app_probe_residual}
\end{equation}
The estimator is fitted from the pairs
$\{(x_p,R_p)\}_{p\in\mathcal P}$.

\paragraph{Group coverage weights.}
Let $m_a=|\mathcal P\cap G_a^Q|$ be the number of probes selected from
query group $a$. Each probe $p\in G_a^Q$ receives weight
\begin{equation}
    w_p=\frac{|G_a^Q|}{m_a}.
    \label{eq:app_probe_weight}
\end{equation}
Each represented group has $m_a>0$ and total raw weight
$\sum_{p\in\mathcal P\cap G_a^Q}w_p=|G_a^Q|$. Selecting more probes
from a group therefore does not increase its total raw weight. Groups
without probes contribute no observations. We normalize the weights to
unit mean to form $W$ as in Section~\ref{sec:residual_repair}.
These deterministic coverage weights do not correct for random sampling.

\subsection{Weighted Affine Fitting and Residual Prediction}
\label{app:regularized_fit}

For the probe statistics we normalize coverage weights to unit sum.
\begin{equation}
    \alpha_p
    =
    \frac{w_p}{\sum_{t\in\mathcal P}w_t}.
    \label{eq:app_normalized_weights}
\end{equation}
The weighted feature and residual means are
\begin{equation}
    \mu_x
    =
    \sum_{p\in\mathcal P}\alpha_p x_p,
    \qquad
    \mu_R
    =
    \sum_{p\in\mathcal P}\alpha_p R_p.
    \label{eq:app_weighted_means}
\end{equation}
Define the elementwise feature scale
\begin{equation}
    s_x
    =
    \left[
        \sum_{p\in\mathcal P}
        \alpha_p(x_p-\mu_x)^{\odot 2}
        +\epsilon
    \right]^{1/2},
    \label{eq:app_feature_scale}
\end{equation}
where $\odot 2$ denotes elementwise squaring. The standardized feature and
centered residual are
\begin{equation}
    \bar x_p=(x_p-\mu_x)\oslash s_x,
    \qquad
    \bar R_p=R_p-\mu_R,
    \label{eq:app_standardized_probe_data}
\end{equation}
where $\oslash$ denotes elementwise division. Stacking the probe rows gives
$\bar X_{\mathcal P}\in\mathbb R^{M\times d}$ and
$\bar R_{\mathcal P}\in\mathbb R^{M\times d}$.

For the regression objective we normalize coverage weights to unit mean.
\begin{equation}
    \widetilde w_p
    =
    \frac{Mw_p}{\sum_{t\in\mathcal P}w_t},
    \qquad
    W
    =
    \operatorname{diag}
    \left(\{\widetilde w_p\}_{p\in\mathcal P}\right).
    \label{eq:app_regression_weights}
\end{equation}
This preserves relative weights and removes their common scale from the
balance between the fitting loss and ridge penalty.

\paragraph{Weighted ridge fit.}
We fit the centered residual from the standardized sparse output.
\begin{equation}
    B_\lambda
    =
    \arg\min_B
    \left\|
        W^{1/2}
        \left(
            \bar R_{\mathcal P}-\bar X_{\mathcal P}B
        \right)
    \right\|_F^2
    +
    \lambda\|B\|_F^2.
    \label{eq:app_weighted_ridge}
\end{equation}
Its closed form is
\begin{equation}
    B_\lambda
    =
    \left(
        \bar X_{\mathcal P}^\top W\bar X_{\mathcal P}
        +\lambda I
    \right)^{-1}
    \bar X_{\mathcal P}^\top W\bar R_{\mathcal P}.
    \label{eq:app_ridge_closed_form}
\end{equation}
The implementation solves the regularized normal equations with a linear
solver rather than explicitly forming the inverse.

\paragraph{Residual prediction.}
For an unprobed row $i$ we use the same probe statistics to obtain
$\bar x_i=(x_i-\mu_x)\oslash s_x$. The predicted residual is
\begin{equation}
    \widehat R_i
    =
    \mu_R+\bar x_iB_\lambda.
    \label{eq:app_residual_prediction}
\end{equation}
The centered prediction lies in the span of the centered probe residuals.
The default estimator uses no output truncation or correction norm cap.
It requires neither a residual SVD nor an output projection. We evaluate
output rank truncation only in Appendix~\ref{app:probe_sensitivity}.
The final output is
\begin{equation}
    \widehat O_i
    =
    \begin{cases}
        O_i^{\mathrm{dense}}, & i\in\mathcal P,\\[1mm]
        O_i^{\mathrm{sp}}+\widehat R_i, & i\notin\mathcal P.
    \end{cases}
    \label{eq:app_final_output}
\end{equation}

\section{Response Coupled Partitioning and Sparse Execution}
\label{app:partition_execution}

\subsection{Key Response Coordinates for Paired K/V Groups}
\label{app:key_response_coordinates}

For each K/V head we form $Q_s\in\mathbb R^{M_s\times d}$ from query rows
at deterministic evenly spaced token positions in the current call.
The key response metric is
\begin{equation}
    M_K=\frac{1}{M_s}Q_s^\top Q_s.
    \label{eq:app_key_metric}
\end{equation}
Let $\Delta k=k_j-k_\ell$ for two keys and let $q_i$ denote row $i$ of
$Q_s$. Their average squared response difference is
\begin{equation}
\begin{aligned}
    \frac{1}{M_s}\sum_{i=1}^{M_s}(q_i\Delta k^\top)^2
    &=
    \Delta k
    \left(\frac{1}{M_s}\sum_{i=1}^{M_s}q_i^\top q_i\right)
    \Delta k^\top \\
    &=
    \Delta k
    \left(\frac{Q_s^\top Q_s}{M_s}\right)
    \Delta k^\top
    =
    \Delta k M_K\Delta k^\top.
\end{aligned}
    \label{eq:app_key_distance}
\end{equation}
The metric therefore compares responses to the sampled queries before
softmax. Dividing this quantity by $d$ gives the average squared difference
between logits scaled by $1/\sqrt d$. We retain the leading $r_K$
eigendirections of $M_K$ and normalize their eigenvalues to unit mean.
Scaling each direction by the square root of its normalized eigenvalue
gives $F_K\in\mathbb R^{d\times r_K}$. The key response coordinate is
\begin{equation}
    \phi_K(k_j)=k_jF_K\in\mathbb R^{r_K}.
    \label{eq:app_key_response_coordinate}
\end{equation}
Euclidean distance in these coordinates is equivalent to a low-rank
Mahalanobis-type distance in the original key space, with response-induced
metric $F_KF_K^\top$.
Clustering $\phi_K(k_j)$ with $k$ means produces paired K/V groups.
Each value $v_j$ inherits the assignment of its paired key $k_j$.
Values do not enter the clustering features.

\subsection{Query Response Coordinates from K/V Groups}
\label{app:query_response_coordinates}

Here $C_k$ counts nonempty K/V groups and may be smaller than the requested
cluster count. We exclude empty groups from the centroid mean and the
query response metric. We compute each centroid $\bar k_b$ in the original
key space and subtract the uniform centroid mean $\mu_K$. The centered
centroids form the rows of $\widetilde K\in\mathbb R^{C_k\times d}$.
For query $q_i$ we have
\begin{equation}
    \frac{q_i(\bar k_b-\mu_K)^\top}{\sqrt d}
    =
    \frac{q_i\bar k_b^\top}{\sqrt d}
    -
    \frac{q_i\mu_K^\top}{\sqrt d}.
    \label{eq:app_query_logit_centering}
\end{equation}
The subtracted term is shared by every group. Centering therefore preserves
relative logits and their softmax over groups. The query response metric is
\begin{equation}
    M_Q
    =
    \frac{\widetilde K^\top\widetilde K}{C_kd}.
    \label{eq:app_query_metric}
\end{equation}
For two queries $q_i$ and $q_j$ we obtain
\begin{equation}
    (q_i-q_j)M_Q(q_i-q_j)^\top
    =
    \frac{1}{C_kd}
    \left\|\widetilde K(q_i-q_j)^\top\right\|_2^2.
    \label{eq:app_query_distance}
\end{equation}
Distance under $M_Q$ is the average squared difference between the
centered logit profiles over groups. Each nonempty group receives equal
weight regardless of its token count.

Let $F_Q\in\mathbb R^{d\times r_Q}$ be obtained from the leading $r_Q$
eigendirections of $M_Q$ using the same eigenvalue weighting as above. The
normalized query response coordinate is
\begin{equation}
\begin{aligned}
    \phi_Q(q_i)
    &= \operatorname{RMSNorm}(q_iF_Q) \\
    &=
    \frac{q_iF_Q}
    {\max\!\left(
        \sqrt{\|q_iF_Q\|_2^2/r_Q},
        \epsilon
    \right)}
    \in\mathbb R^{r_Q}.
\end{aligned}
    \label{eq:app_query_response_coordinate}
\end{equation}
Before RMS normalization, Euclidean distance in the projected coordinates
defines a low-rank Mahalanobis-type metric in query space. RMS normalization
then makes clustering depend primarily on response direction.
The floor $\epsilon=10^{-6}$ stabilizes the denominator. RMS normalization
has no learned gain or bias.
It does not preserve softmax concentration. We retain the original
queries for routing and attention.
Clustering $\phi_Q(q_i)$ gives the query partition. These coordinates
also stratify probe selection but do not enter residual regression.
The construction proceeds from K/V groups to query groups without
alternating refinement.

\subsection{Routing Score and Executed Pair Budget}
\label{app:routing_budget}

Values enter the routing score but not the clustering features.
Let $\bar q_a$ be the query group centroid and let $\bar k_b$ and
$\bar v_b$ be the key and value centroids of $G_b^{KV}$ in their original
spaces. We use the SVG-EAR value error score~\citep{zhou2026svg}.
\begin{equation}
    s_{ab}
    =
    \frac{1}{|G_b^{KV}|}
    \sum_{j\in G_b^{KV}}
    \left\|
        e^{\bar q_a\bar k_b^\top/\sqrt d}\bar v_b
        -
        e^{\bar q_a k_j^\top/\sqrt d}v_j
    \right\|_2^2.
    \label{eq:app_routing_score}
\end{equation}
For each query group we rank nonempty K/V groups by decreasing score.
We retain the shortest prefix that reaches the common key token budget
per query from Section~\ref{sec:kernel_budget}. Each selected cell costs
$|G_a^Q||G_b^{KV}|$ exact interactions. We evaluate selected cells exactly
and renormalize attention over their keys. Centroids determine routing
scores but do not supply approximate output contributions.

After routed sparse attention we evaluate probes against all keys.
Writing the probe fraction as $m=M/N_q$ gives total executed pair density
\begin{equation}
    \rho_{\mathrm{exec}}
    =
    \rho_{\mathrm{route}}+m.
    \label{eq:app_executed_density}
\end{equation}
For target $\rho^\star$ the router reserves probe cost and uses routed
budget $\rho_{\mathrm{route}}^\star=\rho^\star-m$. Selecting whole cells
can make realized density differ from the target. We report realized
density for each configuration.

\section{Implementation and Evaluation Details}
\label{app:implementation_protocol}

\subsection{GPU Implementation}

We cluster on the GPU and fuse query response projection with RMS
normalization. Initial $k$ means centers are sampled uniformly with
replacement from feature rows. Later calls initialize from the previous
centroids in the original space projected into the current response
coordinates. An FP32 eigendecomposition initializes the metric factors.
Later calls update them with two subspace iterations initialized from
the previous factors and using Cholesky factorization.
At the evaluated production shape the fused query projection reduces
latency from $1.6170$ ms to $0.5064$ ms with relative feature error
$3.68\times10^{-4}$ against the reference implementation. Small batched
GPU linear algebra operations fit and apply the residual correction.
The default estimator uses no residual SVD or output subspace projection.

\subsection{Shared Hyperparameters}

All experiments use one NVIDIA H100 GPU with BF16 model activations.
Metric construction and centroid statistics use FP32 as does factorization.
The optimized query projection uses TF32 multiplication with FP32
accumulation and normalization. Models share the method settings except
for the total executed pair target.

\begin{table}[H]
\centering
\small
\caption{Method configuration used in the evaluated models.}
\label{tab:aux_hyperparameters}
\setlength{\tabcolsep}{6pt}
\begin{tabular}{@{}lc@{}}
\toprule
\textbf{Parameter} & \textbf{Value} \\
\midrule
Requested query clusters $C_q$ & $300$ \\
Requested K/V clusters $C_k$ & $1000$ \\
Metric query sampling budget per query head & $2048$ \\
Initial / subsequent $k$ means iteration limits & $25 / 2$ \\
$k$ means convergence tolerance & $10^{-4}$ \\
Key response rank $r_K$ & $48$ \\
Query response rank $r_Q$ & $64$ \\
Exact probes per query head $M$ & $64$ \\
Output subspace truncation & None \\
Ridge coefficient $\lambda$ & $1$ \\
Target $\rho_{\mathrm{exec}}$ for HunyuanVideo / Wan2.2 / Cosmos-Predict2.5
& $0.22$ \\
Target $\rho_{\mathrm{exec}}$ for Cosmos3-Nano & $0.26$ \\
\bottomrule
\end{tabular}
\end{table}

For GQA we concatenate query rows from the \(G\) heads sharing a K/V head.
We select up to \(2048G\) evenly spaced rows to construct the shared key
response metric and use all rows when fewer are available.

\begin{table}[H]
\centering
\small
\caption{Probe fractions for the evaluated sequence lengths with $M=64$.}
\label{tab:aux_probe_fractions}
\setlength{\tabcolsep}{7pt}
\begin{tabular}{@{}lrr@{}}
\toprule
\textbf{Model} & $\boldsymbol{N_q}$ & $\boldsymbol{M/N_q}$ \\
\midrule
HunyuanVideo & $118{,}800$ & $0.054\%$ \\
Wan2.2 & $75{,}600$ & $0.085\%$ \\
Cosmos-Predict2.5 & $84{,}480$ & $0.076\%$ \\
Cosmos3-Nano & $44{,}160$ & $0.145\%$ \\
\bottomrule
\end{tabular}
\end{table}

\subsection{Timing Protocol}

Dense and sparse runs use the same checkpoints and conditioning inputs.
Random seeds and sampling schedules also match. We hold inference steps
and guidance settings fixed along with resolution and frame count.
Executed pair density counts all exact interactions between queries and
keys including probes. Reported latency includes response coordinate
construction and clustering together with routing and permutation.
It also includes sparse attention and exact probes as well as residual
fitting and correction followed by output restoration.

\section{Additional Ablations and Sensitivity Analyses}
\label{app:ablations}

\subsection{Matched Retained Mass and Output Error}
\label{app:matched_mass_error}

\begin{figure}[!ht]
    \centering
    \includegraphics[width=\textwidth]{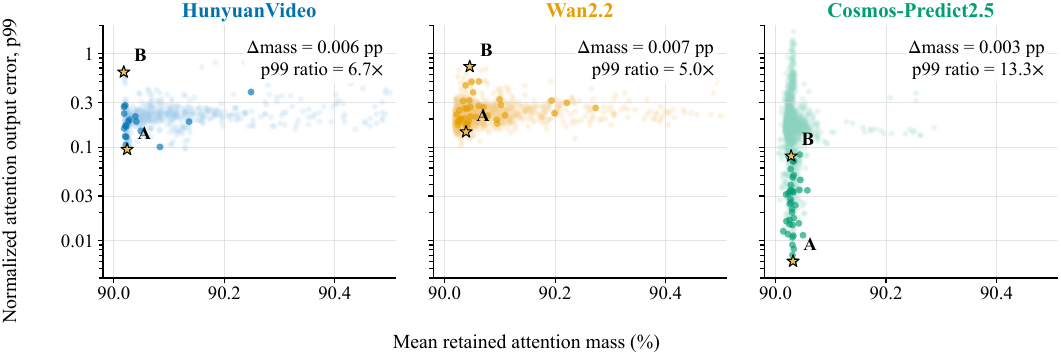}
    \caption{\textbf{Matched retained mass across models.}
    Dark points are heads from the same input, layer, and generation step
    as Cases A and B. Stars mark two heads with nearly identical retained
    mass. The logarithmic y axis shows multiplicative differences in p99
    normalized output error.}
    \label{fig:matched_mass_other_models}
\end{figure}

Figure~\ref{fig:matched_mass_other_models} extends O2 to HunyuanVideo,
Wan2.2, and Cosmos-Predict2.5. Each point summarizes one attention head
at a fixed input, layer, and generation step. Its coordinates are the
mean retained attention mass and p99 normalized output error across queries.

Within each panel, Cases A and B differ by at most $0.007$ percentage
points in retained mass, while their p99 errors differ by
$5.0\times$ to $13.3\times$.

\subsection{Query RMS Normalization}
\label{app:query_rms_ablation}

We compare RMS normalization per query with a shared RMS scale per head
and with RMS normalization per query plus a log magnitude coordinate.
The shared scale preserves relative query magnitudes. For the added
coordinate we center and standardize the log RMS magnitude across query
rows and multiply it by $\sqrt{r_Q}$ with magnitude weight one.

All variants share K/V groups and query projections as well as clustering
seeds and evaluation rows. We use 32 captured calls per model with all
query heads and 2048 evenly spaced evaluation queries per head.
Requested group counts are $C_q=300$ and $C_k=1000$ with coordinate ranks
$r_K=48$ and $r_Q=64$. We load activations in BF16 and replay attention
in FP32. Clustering starts from fresh centers with at most 25 iterations.
Each variant uses the same routing rule with a budget calibrated toward
$22\%$ executed pair density. We pool query errors across calls and heads
for renormalized hard drop without residual repair.

\begin{table}[t]
\centering
\small
\caption{\textbf{Query RMS normalization ablation.}
Mean normalized attention output error
$\lVert O_i^{\mathrm{sp}}-O_i^{\mathrm{dense}}\rVert_2/
(\lVert O_i^{\mathrm{dense}}\rVert_2+10^{-8})$. Lower is better.
Bold entries mark the lowest error in each row.}
\label{tab:query_rms_ablation}
\setlength{\tabcolsep}{4pt}
\begin{tabular*}{\textwidth}{@{\extracolsep{\fill}}lrrr@{}}
\toprule
\textbf{Model} & \shortstack{Per query\\RMS} & \shortstack{Global\\RMS} & \shortstack{Per query RMS\\+ log magnitude} \\
\midrule
HunyuanVideo & \textbf{0.073} & 0.074 & 0.083 \\
Wan2.2 & \textbf{0.108} & 0.109 & 0.166 \\
Cosmos-Predict2.5 & \textbf{0.761} & 0.762 & 0.777 \\
Cosmos3-Nano & \textbf{0.331} & 0.332 & 0.337 \\
\bottomrule
\end{tabular*}
\end{table}

RMS normalization per query gives the lowest mean error across all four
models in Table~\ref{tab:query_rms_ablation} with a small improvement over
global RMS. The table includes all cases even when discrete routing
budgets prevent the target density from being reached.
For Cosmos-Predict2.5 we also evaluate 1247 of the 1280 cases where all
variants lie within $0.2$ percentage points of the target and of each
other. The ordering is unchanged. Mean error is $0.8227$ for RMS per query
and $0.8247$ for global RMS. Adding log magnitude gives $0.8423$.

\subsection{Probe and Estimator Sensitivity}
\label{app:probe_sensitivity}

Table~\ref{tab:probe_rank_sensitivity} evaluates the affine estimator
using sparse outputs as its only features. It covers 32 captured calls
per model and all query heads for a total of 4352 cases. The reference
uses $M=64$ and $\lambda=1$ without output truncation.
For the rank ablation we project the centered ridge prediction onto
the leading $r$ right singular vectors of $W^{1/2}\bar R_{\mathcal P}$.
The residual mean $\mu_R$ is added separately without truncation.
We use response coupled partitions and nested probes drawn from full
query population. Each head has 2048 deterministic evaluation candidates.
We exclude all 128 candidate probes to keep same unprobed evaluation
rows across settings and pool errors across captures and heads.

We load BF16 activations and replay attention in FP32. Regression and
residual subspace fitting use FP64. Feature standardization uses
$\sqrt{\text{weighted variance}+10^{-6}}$.
Sparse base remains fixed for each model at the route calibrated
to a $22\%$ total executed pair target with $M=64$. Varying $M$ therefore
changes total cost. 

\begin{table}[t]
\centering
\small
\caption{
\textbf{Sensitivity to probe count and estimator settings.}
Entries report mean normalized attention output error on the same unprobed
query rows across settings. Lower is better. Each panel varies one parameter
while holding the others fixed.
}
\label{tab:probe_rank_sensitivity}
\setlength{\tabcolsep}{4pt}
\begin{tabular*}{\textwidth}{@{\extracolsep{\fill}}lrrrr@{}}
\toprule
\multicolumn{5}{@{}l}{\textit{(a) Probe count with fixed $\lambda=1$ and no output truncation}} \\
\midrule
\textbf{Model} & $M=16$ & $M=32$ & $\boldsymbol{M=64}$ & $M=128$ \\
\midrule
HunyuanVideo & 0.037 & 0.034 & \textbf{0.030} & 0.027 \\
Wan2.2 & 0.080 & 0.071 & \textbf{0.063} & 0.057 \\
Cosmos-Predict2.5 & 0.118 & 0.092 & \textbf{0.075} & 0.065 \\
Cosmos3-Nano & 0.097 & 0.082 & \textbf{0.076} & 0.074 \\
\bottomrule
\end{tabular*}

\vspace{0.8em}
\begin{tabular*}{\textwidth}{@{\extracolsep{\fill}}lrrrrr@{}}
\toprule
\multicolumn{6}{@{}l}{\textit{(b) Output rank truncation with fixed $M=64$ and $\lambda=1$}} \\
\midrule
\textbf{Model} & $r=4$ & $r=8$ & $r=16$ & $r=32$ & \textbf{No truncation} \\
\midrule
HunyuanVideo & 0.034 & 0.032 & 0.031 & 0.031 & \textbf{0.030} \\
Wan2.2 & 0.072 & 0.068 & 0.066 & 0.065 & \textbf{0.063} \\
Cosmos-Predict2.5 & 0.113 & 0.089 & 0.079 & 0.077 & \textbf{0.075} \\
Cosmos3-Nano & 0.089 & 0.081 & 0.078 & 0.077 & \textbf{0.076} \\
\bottomrule
\end{tabular*}

\vspace{0.8em}
\begin{tabular*}{\textwidth}{@{\extracolsep{\fill}}lrrrr@{}}
\toprule
\multicolumn{5}{@{}l}{\textit{(c) Ridge strength with fixed $M=64$ and no output truncation}} \\
\midrule
\textbf{Model} & $\lambda=0.001$ & $\lambda=0.01$ & $\lambda=0.1$ & $\boldsymbol{\lambda=1}$ \\
\midrule
HunyuanVideo & 0.039 & 0.038 & 0.034 & \textbf{0.030} \\
Wan2.2 & 0.086 & 0.080 & 0.072 & \textbf{0.063} \\
Cosmos-Predict2.5 & 0.095 & 0.086 & 0.080 & \textbf{0.075} \\
Cosmos3-Nano & 0.095 & 0.090 & 0.084 & \textbf{0.076} \\
\bottomrule
\end{tabular*}

\vspace{0.4em}
\parbox{\textwidth}{\footnotesize
Bold headers and entries mark the reference configuration. Requested ranks in panel (b) are
capped at the observed centered probe residual rank of at most $M-1$. No truncation denotes ridge regression without output subspace
truncation. Increasing $M$ in panel (a) also increases total executed pair density.}
\end{table}

\subsection{Correction Family Comparison}
\label{app:correction_family}

Table~\ref{tab:probe_estimator_ablation} compares correction by the
weighted mean residual with three affine variants. Each affine variant
predicts $\mu_R+\bar x_i B$. The scalar variant uses $B=aI$ and the
diagonal variant uses $B=\operatorname{diag}(b)$. Full weighted ridge
leaves $B$ unrestricted. All three minimize the same weighted squared
error with penalty $\lambda\lVert B\rVert_F^2$ and $\lambda=1$.
The scalar penalty is therefore $\lambda d a^2$. We do not penalize
the residual mean or truncate the output.

All variants share sparse outputs and unprobed evaluation rows.
They use the same $M=64$ probes with identical coverage weights and
feature standardization. Evaluation rows and numerical precision
follow Appendix~\ref{app:probe_sensitivity}.

\subsection{Oracle Affine Residual Representability}
\label{app:oracle_affine_representability}

We test O3 by comparing semantic and response coupled partitions using
the same affine features $X_g=[\mathbf{1},O_g^{\mathrm{sp}}]$.
For each captured call and head we compute dense and sparse outputs on
2048 evenly spaced queries and fit the predictor to all of them.
This oracle fit measures representability on the fitted rows before
estimating a correction from a limited probe set.

\begin{figure}[!ht]
    \centering
    \includegraphics[width=\textwidth]{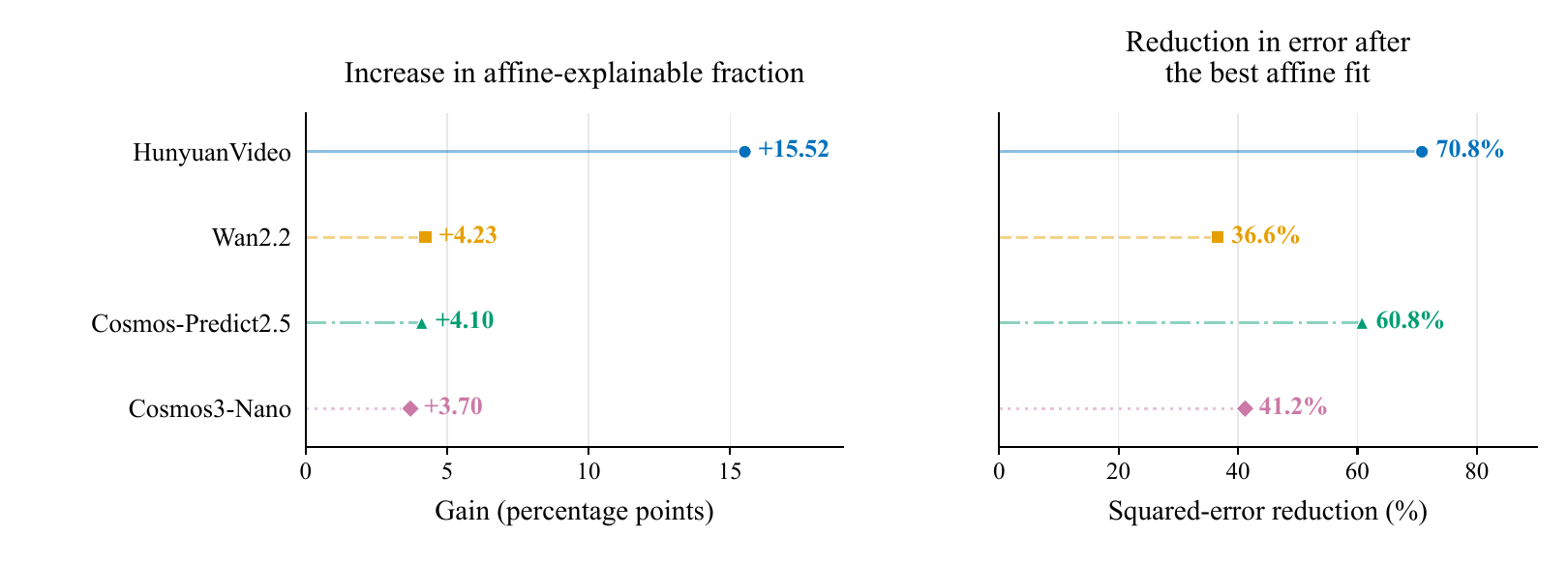}
    \caption{Oracle affine residual representability under response coupled
    versus semantic partitioning. The left panel shows the increase in
    the fraction of residual squared norm explained by the best affine fit.
    The right panel shows the reduction in squared error remaining after
    fitting. We fit all 2048 evaluated queries per head and sum energies
    across cases before computing ratios. Cosmos-Predict2.5 uses the
    subset matched by density.}
    \label{fig:oracle_affine_representability}
\end{figure}

We use 32 captured calls per model with all query heads. This gives
768 cases for HunyuanVideo and 1280 for Wan2.2. Cosmos-Predict2.5 has
1280 cases and Cosmos3-Nano has 1024. Both partitions request $C_q=300$
and $C_k=1000$ and use the same routing rule with renormalized hard drop.
Routing targets $22\%$ total executed pair density and reserves the cost
of 64 exact probes as in the repair configuration. This reservation
does not restrict the oracle fit.
The Cosmos-Predict2.5 figure uses 1248 cases matched by density and excludes
cases where discrete routing budgets prevent matching. The explained
fraction increases by $4.10$ percentage points in this subset and by
$4.08$ across all 1280 cases. The reduction in remaining squared error
rounds to $60.8\%$ for both populations. Other models use all cases.

We load activations in BF16 and compute attention outputs in FP32.
The oracle projection uses FP64 and an SVD of centered sparse outputs
with scaled columns. The relative singular value tolerance is $10^{-10}$.
The residual mean supplies the affine intercept.

Figure~\ref{fig:oracle_affine_representability} reports the gain in
explained residual energy in percentage points and the reduction in
remaining squared error relative to semantic partitioning. We sum
squared errors across the same cases before computing fractions and reductions.

\paragraph{Interpretation.}
Response coupled partitioning increases the fraction explained by the
affine fit and reduces remaining squared error across all four models.
The remaining error depends on both the initial residual magnitude
and the fraction explained. These results support the capacity of the
affine correction. Table~\ref{tab:partition_repair_ablation} separately
evaluates prediction on unprobed rows after fitting from a small probe set.
\FloatBarrier

\subsection{Further Discussion}
\label{sec:further_discussion}

We examine repair reliability and alternative estimators with $M=64$ and $\lambda=1$. Comparisons use identical
sparse outputs and unprobed evaluation rows within each case.

\paragraph{Does repair ever worsen individual heads?}
Repair improves aggregate accuracy but can increase error for an
individual head on a particular call. Mean normalized error increases
in 13.05\% of HunyuanVideo cases and 11.41\% of Wan2.2 cases.
The rates are 1.41\% for Cosmos-Predict2.5 and 13.98\% for Cosmos3-Nano.
Each case is one head at a sampled layer and diffusion step.
For Cosmos3-Nano layer 20 the mean error increases by 2.09\% after
pooling evaluated heads and steps. These are rates of worsening
attention error and do not measure generation failures.

\paragraph{Does a small nonlinear predictor improve reconstruction?}
We compare full weighted ridge with a GELU predictor with one hidden
layer of width 32. We train it for 200 Adam steps under five initialization
seeds using the same probes and weights. The nonlinear predictor has
higher mean error on all four models. For Cosmos3-Nano it lowers p99
error from 0.447 to 0.430 but increases mean error by 8.26\%.
Fitting takes 136 to 139\,ms per head in the FP32 H100 benchmark compared
with approximately 0.46\,ms for ridge. These results favor ridge over
the tested nonlinear configuration with 64 probes.

\section{More Qualitative Results}
\label{app:qualitative}

\begin{figure*}
    \centering
    \includegraphics[width=0.99\linewidth]{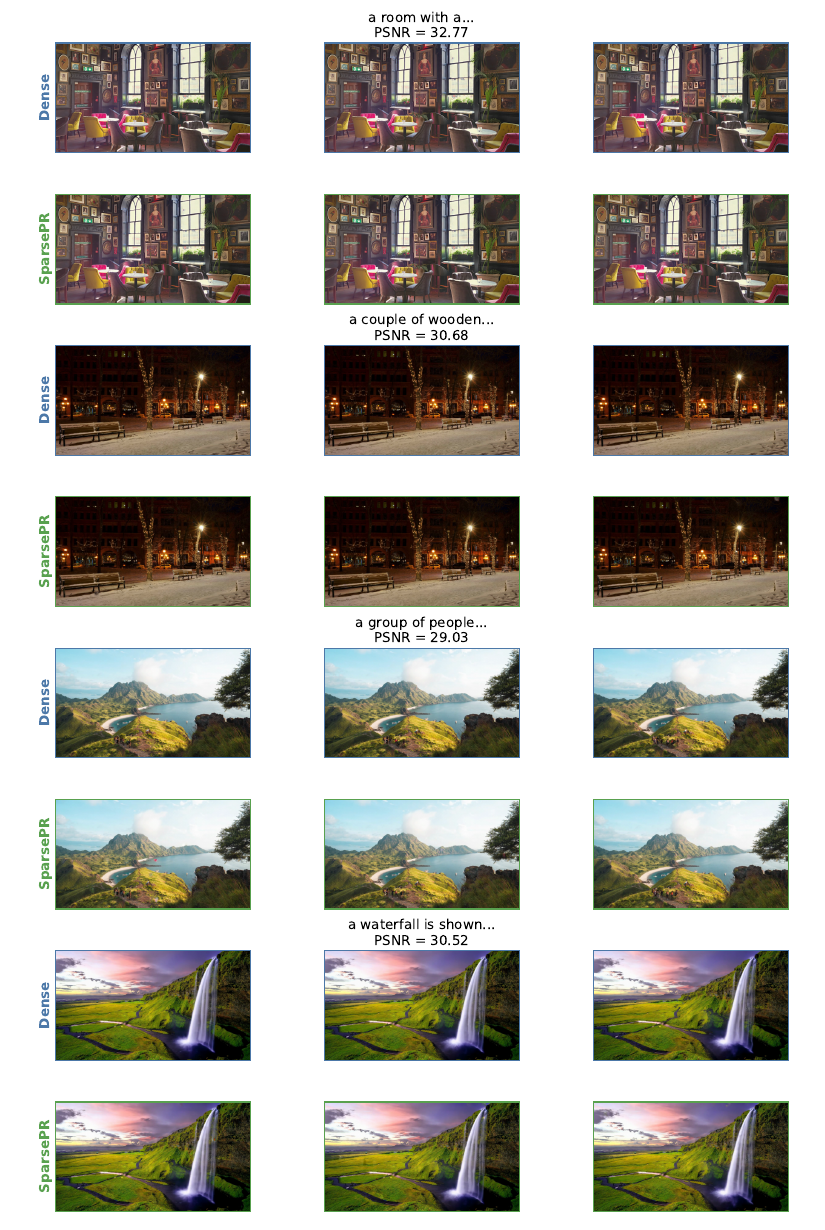}
    \caption{Dense and SparsePR outputs for Cosmos3-Nano-16B on VBench++.}
    \label{fig:qual_cosmos3_vbench}
\end{figure*}

\begin{figure*}
    \centering
    \includegraphics[width=0.99\linewidth]{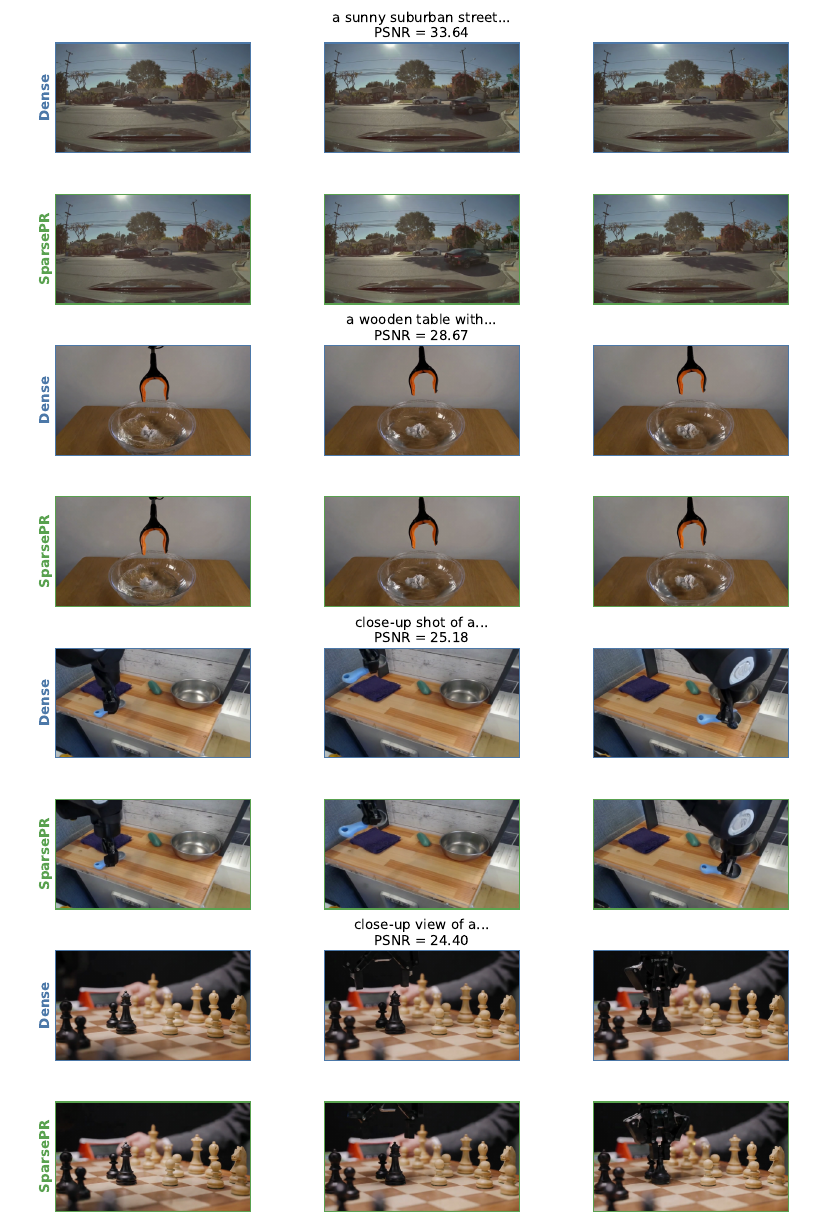}
    \caption{Dense and SparsePR outputs for Cosmos3-Nano-16B on PBench.}
    \label{fig:qual_cosmos3_pbench}
\end{figure*}

\begin{figure*}
    \centering
    \includegraphics[width=0.99\linewidth]{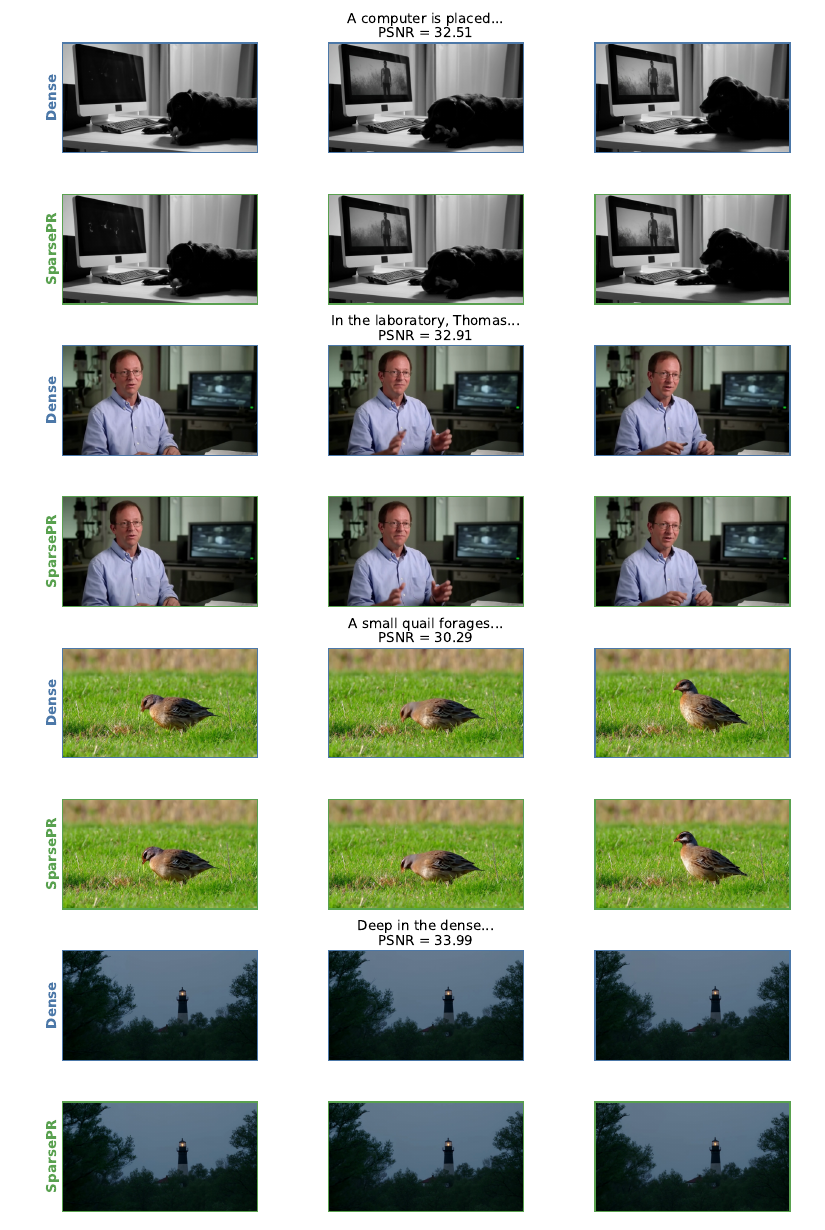}
    \caption{Qualitative comparison on HunyuanVideo-13B.}
    \label{fig:qual_hunyuan}
\end{figure*}

\begin{figure*}
    \centering
    \includegraphics[width=0.99\linewidth]{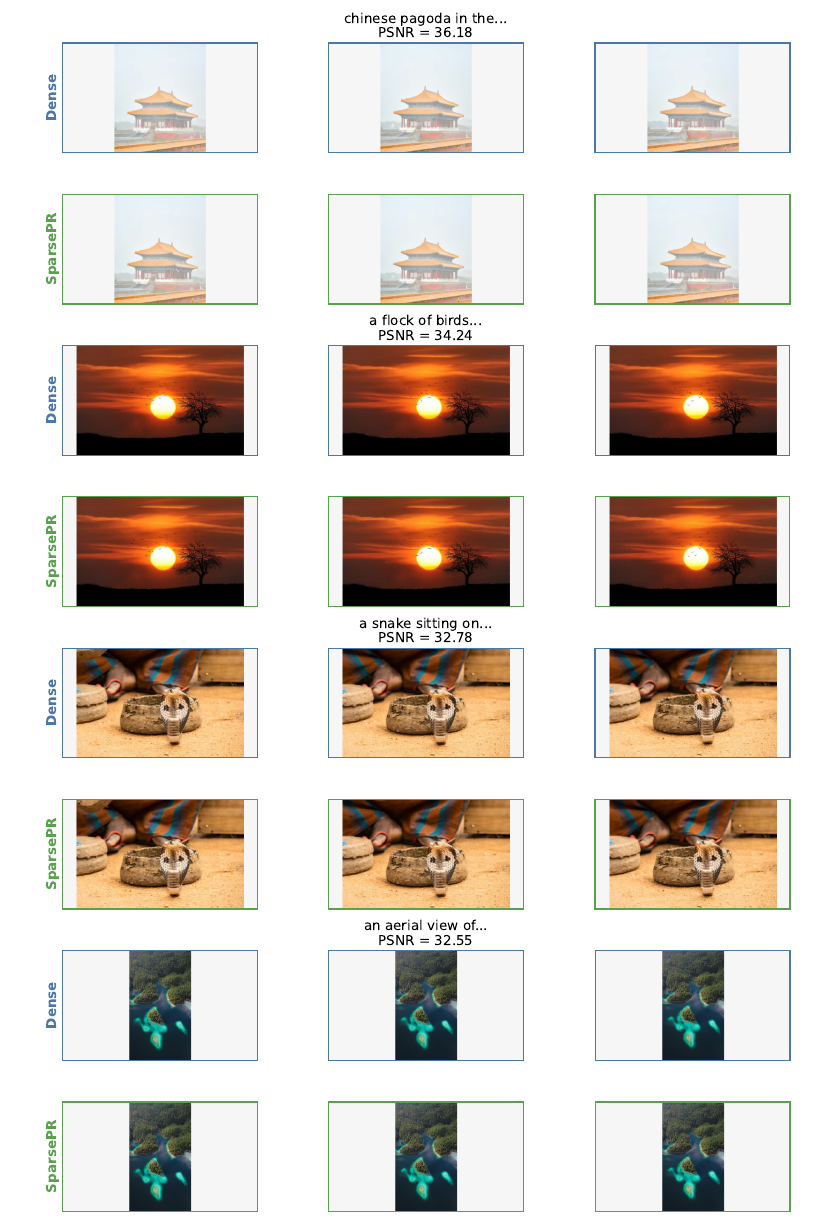}
    \caption{Qualitative comparison on Wan2.2-I2V-A14B.}
    \label{fig:qual_wan22}
\end{figure*}

Figures~\ref{fig:qual_cosmos3_vbench} through~\ref{fig:qual_wan22}
provide additional qualitative comparisons between dense attention and SparsePR.

\end{document}